\documentclass[10pt,twocolumn,letterpaper]{article}
\usepackage[pagenumbers]{cvpr}%
\usepackage{amsmath,amssymb,booktabs,graphicx}
\newcommand{\JECT}{JECT}

\newcommand{\sg}{\operatorname{sg}}

\definecolor{cvprblue}{rgb}{0.21,0.49,0.74}
\usepackage[pagebackref,breaklinks,colorlinks,allcolors=cvprblue]{hyperref}
\def\confName{CVPR}
\def\confYear{2027}
\title{JECT: Addressing the Context Shortcut in\\Joint-Embedding Predictive Learning for Chest CT}
\author{
Hanbin Ko$^{1,2}$ \quad Hong-Jun Choi$^{3}$ \quad Jihun Kim$^{4,5}$ \\
Young-Gon Kim$^{4,6,7}$ \quad Dongheon Lee$^{8,9}$ \quad Chang Min Park$^{8,10}$ \\[6pt]
{\footnotesize
\begin{tabular}{@{}c@{}}
$^{1}$Interdisciplinary Program in Bioengineering, Seoul National University Graduate School\\
$^{2}$Integrated Major in Innovative Medical Science, Seoul National University Graduate School\\
$^{3}$Teamreboott Inc., Busan\\
$^{4}$Department of Transdisciplinary Medicine, Seoul National University Hospital\\
$^{5}$Department of Applied Bioengineering, Graduate School of Convergence Science and Technology,\\ Seoul National University, Suwon\\
$^{6}$Department of Medicine, Seoul National University College of Medicine\\
$^{7}$Healthcare AI Research Institute, Seoul National University Hospital\\
$^{8}$Department of Radiology, Seoul National University Hospital, Seoul National University College of Medicine\\
$^{9}$Interdisciplinary Program in Artificial Intelligence, Seoul National University\\
$^{10}$Institute of Medical and Biological Engineering, Seoul National University Medical Research Center\\[2pt]
All affiliations are in Seoul, Korea, unless otherwise noted.
\end{tabular}}
}
\begin{document}
\maketitle
\begin{abstract}
Joint-embedding predictive architectures learn by predicting features of masked regions from visible context. In chest computed tomography, the findings that decide a diagnosis, such as a small nodule, are often not predictable from the surrounding anatomy. A hidden finding is then best predicted toward normal anatomy, and the prediction loss stops rewarding the encoder for representing it. We call this the \emph{context shortcut}. Conditioning the predictor on the radiology report repairs the prediction, but in our setup the encoder gains little over image-only pretraining: the report supplies what the encoder never had to learn. \JECT{} therefore gives the report two roles: a condition for the predictor and, through masked section-text prediction, a target for the encoder, which must predict the report's anatomical sections from the unmasked scan. A hierarchical three-dimensional encoder that predicts at every stage keeps this clinical content at every scale. In ablations, the text target improves classification and localization, while conditioning adds little to the encoder. Trained from scratch on 173k volumes, the recipe reaches the level of current computed tomography foundation models in classification, retrieval and nodule malignancy, and of task-specific models in lung-cancer risk prediction after fine-tuning. Conditioning still matters: its predictor can be reused, with no extra training, to show where a report sentence acts, changing most at the finding the sentence describes. Code and pretrained weights are available at \url{https://github.com/lukeingawesome/JECT}.
\end{abstract}

\section{Introduction}

\begin{figure}[t]
\centering
\includegraphics[width=\linewidth]{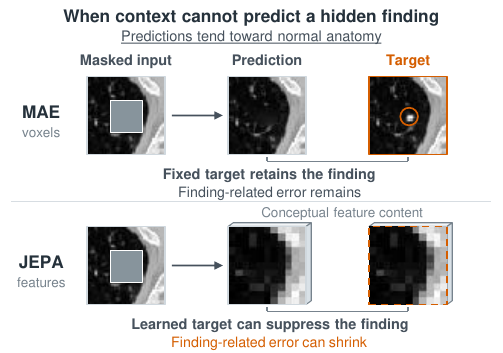}
\caption{\textbf{The context shortcut.} When a finding is hidden, the best prediction is normal anatomy, for pixel (MAE) and feature (JEPA) targets alike. The error that remains cannot be reduced from the context, so it does not teach the encoder the finding. A JEPA target comes from the encoder, which could also lower the error by representing the finding less.}
\label{fig:mechanism}
\end{figure}

\begin{figure*}[t]
\centering
\includegraphics[width=\textwidth]{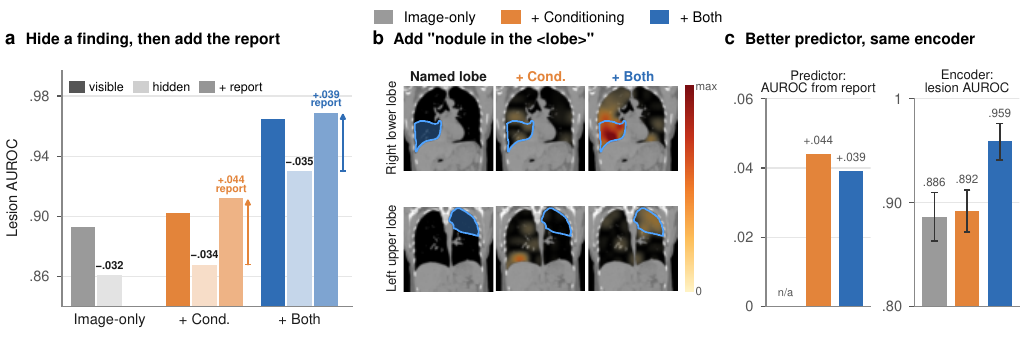}
\caption{\textbf{Conditioning repairs the predictor, not the encoder.} Arms without weak grounding: image-only, +Conditioning, and +Both (conditioning and masked section-text prediction, MSP). (a) Lesion AUROC of the predictor's fill-in for visible and hidden findings, without and with the report (Table~\ref{tab:normal-shift}); bold: drop when hidden. (b) Prediction change after adding ``There is a 2\,cm nodule in the $\langle$lobe$\rangle$''; named lobe in blue. (c) The report raises the predictor's hidden-finding AUROC in both conditioned arms, but MSP raises the encoder's performance.}
\label{fig:clinical-loss}
\end{figure*}

Image encoders are pretrained without labels along three main routes, and all three have reached 3D CT. Masked autoencoders hide parts of an image and reconstruct the missing pixels~\cite{he2022mae}, and have pretrained chest CT transformers on lung-screening scans~\cite{mcconnell2025tangerine}. Contrastive methods align each image with its paired text~\cite{radford2021clip,zhai2023siglip}; CT-CLIP, Merlin and Pillar-0 align CT volumes with their radiology reports~\cite{hamamci2026ctrate,blankemeier2026merlin,agrawal2025pillar0}, and COLIPRI, SPECTRE and FlexiCT combine self-supervised pretraining with report alignment~\cite{wald2026colipri,claessens2025spectre,flexi3d}. Joint-embedding predictive architectures (JEPA) instead predict the features of the hidden parts from the visible ones~\cite{assran2023ijepa,bardes2024vjepa}, so the model need not reproduce low-level detail; JEPA has been trained on more than 100k CT volumes~\cite{trinh2026radjepa3d,drexel2026predictability}.

Masked latent prediction, however, has a \emph{context shortcut}: a route to a low loss that never requires the findings that matter clinically. The findings that decide a chest CT diagnosis, such as a small nodule, are often what the surrounding anatomy does not predict. When such a finding is hidden, the best prediction from the visible context is close to normal anatomy, and nothing in the objective requires the encoder to represent it. Drexel~\etal~\cite{drexel2026predictability} measure this bias toward normal anatomy in the predictor of a CT JEPA, call it \emph{predictability bias}, and leave open how to fix it. The cost reaches the encoder: an error the predictor cannot reduce gives the encoder no signal from that region to learn the finding (Fig.~\ref{fig:mechanism}). On held-out scans with annotated findings, hiding a finding lowers the lesion AUROC of the predictor's fill-in in every model we train (Fig.~\ref{fig:clinical-loss}a).

In our setup, conditioning on the report repairs the predictor but adds little to the encoder. The radiology report describes the findings a mask may hide, so a natural fix is to give it to the predictor as a condition, as TC-JEPA does with captions~\cite{huang2026tcjepa}. On held-out scans, the predictor recovers a hidden finding from the report, yet its encoder separates lesions as well as one pretrained on images alone (0.892 against 0.886; Fig.~\ref{fig:clinical-loss}c). The shortcut has only moved: the report now supplies the finding, which removes the encoder's reason to drop it but also its reason to learn it.

\begin{figure*}[t]
\centering
\includegraphics[width=\textwidth]{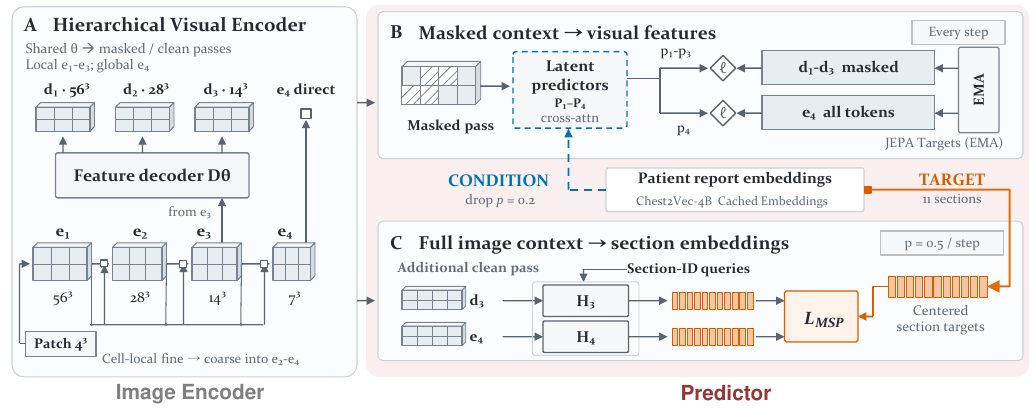}
\caption{\textbf{\JECT{} overview.} (A) Hierarchical encoder with fine-to-coarse attention and a top-down decoder. (B) Predictors at every stage read the masked pass, conditioned on the report's sections, and regress EMA targets. (C) On half of the steps, text heads on the clean pass predict the report's sections (MSP). The report is a condition in (B) and a target in (C).}
\label{fig:architecture}
\end{figure*}

We therefore give the report two roles: a condition for the predictor and a target for the encoder, which gives that reason back (Fig.~\ref{fig:architecture}). \JECT{} (\emph{Joint-Embedding prediction with Clinical Text}) keeps the report as the predictor's condition and adds \emph{masked section-text prediction} (MSP): from the image alone, the encoder must predict what each anatomical section of the report says, given as frozen text embeddings. A section describes its findings whether or not the surrounding anatomy predicts them, so the encoder is rewarded for representing them. A weak-grounding term asks each finding sentence to point to the region it names. To keep this clinical content at every scale, not only in a global summary, the encoder is hierarchical: every stage predicts the features of hidden regions, and connections from finer to coarser stages carry fine detail into the coarse tokens.

MSP differs from other ways of learning from report text in that the report is the encoder's target inside masked feature prediction, alongside the predictor's report condition. VL-JEPA predicts text embeddings~\cite{chen2026vljepa}, Pillar-0 aligns volumes with an embedding of the whole report~\cite{agrawal2025pillar0}, GLoRIA matches words to image regions~\cite{huang2021gloria}, and Jolia's ConQuer contrasts concepts across report sections~\cite{khlaut2026jolia}. This lets us test whether predicting the report gives the encoder what conditioning on it does not.

We work within JEPA and ask how to make it learn the findings that matter clinically. We test each part on held-out scans and with ablations that remove one component at a time, train the recipe at scale, and reuse its predictor to show where report sentences act in the scan (Fig.~\ref{fig:clinical-loss}b). Our contributions are:
\begin{itemize}
\item \textbf{The context shortcut and its cost to the encoder.} Building on the predictor's bias toward normal, we state the encoder's signal for unpredictable findings and test it on held-out scans. We show that conditioning on the report, as in TC-JEPA~\cite{huang2026tcjepa}, repairs the predictor but adds little to the encoder in our setup.
\item \textbf{The report as condition and target.} As the encoder's target, MSP makes the encoder learn the findings, raising classification AUROC by 5 points and nodule localization AUROC by 6 points; conditioning trains the predictor.
\item \textbf{A hierarchical encoder that keeps clinical content at every scale.} Fine-to-coarse connections carry fine detail into the coarse tokens; removing them lowers nodule localization AUROC by 7.5 points.
\item \textbf{A recipe that scales.} Trained from scratch on 173k volumes, it reaches the level of current CT foundation models in classification, report retrieval and nodule malignancy, and of task-specific models in lung-cancer risk prediction after fine-tuning.
\item \textbf{The predictor, reused as a report-grounded readout.} With no extra training, editing a report shows where each sentence acts: deleting a sentence changes the prediction most at the finding it describes, which conditioning without MSP does not achieve.
\end{itemize}

\section{JECT}
\label{sec:method}

\subsection{The context shortcut}
\label{sec:theory}
Masked latent prediction can reach a low loss without learning the hidden finding: the context shortcut. Let $c$ denote the encoded visible context of a volume, $z_h$ the target feature of a hidden region $h$, $r$ the report, and $g$ the predictor. Under an idealized squared-error loss, the best predictor and the loss it leaves are
\begin{equation}
 g^\star(c)=\mathbb E[z_h\mid c],\qquad
 \mathcal L^\star=\mathbb E\,\operatorname{tr}\operatorname{Cov}(z_h\mid c).
\label{eq:optimal}
\end{equation}
These idealized quantities give four arguments with predictions tested in Section~\ref{sec:results}.

\textbf{(A1) Hidden findings are predicted toward normal.} If a finding $F$ occupies $h$ with $\pi=P(F\mid c)$, then $g^\star(c)=\pi\mu_F+(1-\pi)\mu_N$, with $\mu_F,\mu_N$ the expected features with and without it. When the context does not reveal $F$, $\pi$ stays near its prevalence and the prediction moves toward normal anatomy.

\textbf{(A2) The prediction loss gives the encoder no signal for what the visible input cannot predict.} With a stop-gradient target, the gradients of the squared and cosine losses are affine in the target (for cosine, in $z_h/\lVert z_h\rVert$), so in expectation they depend on it only through its mean given the visible input $v$, $\mathbb{E}[z_h \mid v]$. Through this loss, a finding is thus learned only where it is visible and helps predict hidden content. EMA targets could also let the loss fall by representing it less~\cite{littwin2024jepanoisy}; in training we see a ceiling, not a decline (Sec.~\ref{sec:supp-mechanism}).

\textbf{(A3) Conditioning helps the predictor, not necessarily the encoder.} With the report, $\mathbb E\operatorname{Cov}(z_h\mid c,r)\preceq\mathbb E\operatorname{Cov}(z_h\mid c)$: the text makes the finding predictable. The predictor can then lower the loss through $r$, shrinking the error that could have taught the encoder the finding.

\textbf{(A4) A text target restores the encoder's signal.} MSP asks clean image features to predict fixed report-section embeddings, which contain the findings and cannot collapse, so the encoder is rewarded for encoding the findings whether or not the context predicts them. It also pulls image features toward the report's text space, so the predictor's response to an edited report points to places in the image (Sec.~\ref{sec:inference}).

\subsection{Hierarchical encoder}
A four-stage pyramid, trained from random initialization, maps a $224^3$ volume to token grids of $56^3$, $28^3$, $14^3$ and $7^3$ (widths 128-1024), with local attention in $7^3$ windows at stages 1-3 and global attention at stage 4; $e^s$ denotes the output of stage $s$. Because downsampling can average a small structure away, fine-to-coarse connections let each coarse cell attend, before stages 2-4, to the finer tokens inside it from the patch embedding and every earlier stage (query from the coarse feature, four heads of width 32, a learned bias for each position in the cell); the result is added through a zero-initialized projection (Sec.~\ref{sec:supp-method}). A light top-down decoder, as in feature pyramid networks~\cite{lin2017fpn}, refines $e^3$ into $d^3$ and adds each upsampled map to the next finer stage ($d^2$ from $d^3$ and $e^2$, $d^1$ from $d^2$ and $e^1$).

\subsection{Multiscale prediction conditioned on the report}
\label{sec:msp-pred}
\JECT{} hides about half the volume $x$ (cuboids with sides of 4-7 stage-3 cells, replaced by a learned mask token) and, at every stage, predicts the features of the hidden parts, so every resolution must carry what the mask hides~\cite{tian2023itpn}. The online encoder sees the masked volume; an EMA copy ($\xi$) sees the full volume and gives the targets $z^s=d^s_\xi(x)$ for $s\le3$ and $z^4=e^4_\xi(x)$, and the predictor of stage $s$ outputs $p^s$. Stages 3-4 use six-layer transformer predictors and stages 1-2 per-token heads, which get their context through the decoder.

In its first role, the report is a condition for the predictor: each predictor reads it through cross-attention, it is dropped per sample with probability 0.2, and it never enters the encoder. Chest2Vec~\cite{ko2026chest2ve} splits each report into 11 anatomical sections, and its frozen 4B encoder embeds, offline, each section and one sentence sampled from the report. With $\ell(a,b)=1-\cos(a,b)$ and stop-gradient $\sg[\cdot]$, the prediction loss is
\begin{equation}
 \mathcal L_{\mathrm{vis}}=\sum_{s=1}^{3}\frac{\alpha_s}{|M_s|}
 \sum_{j\in M_s}\ell(p_j^s,\sg[z_j^s])
 +\frac{\alpha_4}{N_4}\sum_{j=1}^{N_4}\ell(p_j^4,\sg[z_j^4]),
\label{eq:visual}
\end{equation}
where $M_s$ are the masked tokens at stage $s$, $\alpha=(0.25,0.35,0.40,0.30)$, and stage 4 predicts all $N_4=7^3$ tokens.

\subsection{Masked section-text prediction}
MSP gives the report its second role, a target for the encoder. Text heads at stages 3 and 4 hold one query per section, built from a learned slot and the training-set mean $\mu_k$ of that section, so the query names the section but carries none of the patient's text: the section's content is masked. They attend to $d^3$ and $e^4$ of a clean, unmasked pass and predict $\widehat u^s_{bk}$, the embedding $u_{bk}$ of section $k$ in study $b$ (its normalized 1024-dimensional prefix); the section query also scales each head's output. To reward patient-specific content, the target is the residual $\rho_{bk}=u_{bk}-a_t\mu_k$, normalized to $\widetilde u_{bk}$ and weighted by $w_{bk}=\|\rho_{bk}\|$:
\begin{equation}
 \mathcal L_{\mathrm{MSP}}=
 \frac12\sum_{s\in\{3,4\}}\gamma_s
 \frac{\sum_{b,k}v_{bk}w_{bk}\,
 \ell(\widehat u_{bk}^s,\widetilde u_{bk})}
 {\sum_{b,k}v_{bk}w_{bk}},
\label{eq:msp}
\end{equation}
where $v_{bk}$ marks available sections, $\gamma=(0.5,1)$ and $a_t$ ramps from 0 to 1. The targets come from a frozen text encoder, so the loss needs no negatives.

\subsection{Training objective}
The objective adds the text targets to JEPA training, $\mathcal L_{\mathrm{JEPA}}=\mathcal L_{\mathrm{vis}}+d_t\mathcal L_{\mathrm{dense}}+0.02\,\mathcal L_{\mathrm{SIGReg}}$:
\begin{equation}
 \mathcal L=\mathcal L_{\mathrm{JEPA}}
 +b_t\tau_t\big(0.5\,\mathcal L_{\mathrm{MSP}}+0.5\,\mathcal L_{\mathrm{WG}}\big),
\label{eq:total}
\end{equation}
with the dense loss of V-JEPA~2.1~\cite{mur2026vjepa21} on visible tokens near the mask and SIGReg~\cite{balestriero2025lejepa} against collapse. Weak grounding (WG) asks each finding sentence to point, on $d^2$, to the anatomical region it names (BCE and Dice against a TotalSegmentator~\cite{wasserthal2023totalsegmentator} mask of that region). Half of the steps run the clean pass ($b_t\sim\operatorname{Bernoulli}(0.5)$), and $d_t$ and $\tau_t$ are warm-up ramps (training details in Sec.~\ref{sec:supp-method}).

\subsection{Inference and the report-grounded readout}
\label{sec:inference}
Classification and localization use the unmasked EMA encoder with no report; retrieval adds an alignment stage. The trained predictor gives a second, report-grounded readout: where a report sentence acts. Over eight seeded masks per scan, the residual between the stage-3 prediction and its target $z_j$ for cell $j$ under report $r$ is
\begin{equation}
 R_j(r)=\frac{1}{|\mathcal M_j|}\sum_{m\in\mathcal M_j}
 \left[1-\cos\!\big(p_j^{(m)}(r),z_j\big)\right],
\end{equation}
with $\mathcal M_j$ the masks that hide $j$. Editing the report, for example re-embedding a section without one sentence, changes $R_j$ by $A_j=R_j(r_{\text{edited}})-R_j(r)$; cells whose prediction relied on that text change most. It needs no extra training or labels; Section~\ref{sec:attribution} tests it on lesion annotations.

\section{Experimental Setup}
\label{sec:setup}
\paragraph{Pretraining.}
Bridge-scale (ablation) experiments train on the 23,476 CT-RATE~\cite{hamamci2026ctrate} training volumes for 50 epochs, with microbatch 8 and four-step accumulation (effective batch 32). Each ablation changes one component of the complete recipe. A fixed per-step seed makes mask sampling, report dropout and the clean-pass draw identical across runs at every step. Foundation-scale pretraining uses 173,185 de-identified chest CT volumes from two institutions, including the CT-RATE training studies, on two devices with the same effective batch.

\paragraph{Readouts.}
Every run is evaluated with frozen features and no report entering the image encoder, with four readouts that ask different questions. \emph{Classification} (which findings a scan contains): linear probes on a 1,544-scan CT-RATE validation cohort for the Chest-44~\cite{ko2026chest2ve} and CT-RATE-18~\cite{hamamci2026ctrate} label sets, macro AUROC of the stage-4 token mean.  \emph{Nodule localization} (where a small finding is): logistic readouts on 24\,mm stage-3 decoder cells for LUNA16~\cite{setio2017luna16,armato2011lidc,clark2013tcia}, with cell AUROC and a cell-based CPM (mean sensitivity at 0.125-8 false positives per scan). \emph{Nodule malignancy} (how a nodule at a known location is graded): LUNA25~\cite{peeters2026luna25}, five patient-grouped folds. \emph{Retrieval} (whether a scan identifies its own report): a pooler and projection trained on the frozen encoder, with recall on the 1,564-scan CT-RATE validation gallery. Released models are evaluated from their weights under the same protocols. Cohort counts are in Sec.~\ref{sec:supp-data}. All comparisons use paired patient-bootstrap 95\% intervals; a difference is \emph{separated} when its interval excludes zero (Sec.~\ref{sec:supp-stats}).

\section{Results}
\label{sec:results}
\subsection{Testing the context shortcut}
\label{sec:mechanism}
\begin{table}[t]
\centering\footnotesize
\caption{Hiding a finding on held-out ReXGroundingCT~\cite{baharoon2025rexgroundingct} cases (733 findings). Prob.\ shift: change in finding probability when hidden. Restored: share of the AUROC drop the report brings back. Own $-$ other: how much closer the patient's own report, rather than another's, moves the fill-in to the lesion. Encoder: lesion AUROC of the encoder's own features.}
\label{tab:normal-shift}
\setlength{\tabcolsep}{3pt}
\resizebox{\columnwidth}{!}{\begin{tabular}{lccccccc}
\toprule
& \multicolumn{3}{c}{Hiding (A1)} & \multicolumn{3}{c}{Adding the report} & Encoder\\
\cmidrule(lr){2-4}\cmidrule(lr){5-7}\cmidrule(lr){8-8}
Arm & Visible & Hidden & Prob.\ shift & AUROC & Restored & Own $-$ other & (A2)\\
\midrule
Image-only & .893 & .861 & $-$.050 & \multicolumn{3}{c}{\emph{no report input}} & .886\\
$+$Conditioning & .902 & .868 & $-$.124 & .912 & 129\% & .001 & .892\\
$+$Both & .965 & .930 & $-$.168 & .969 & 111\% & .031 & .959\\
\bottomrule
\end{tabular}}
\end{table}

In every model, a hidden finding is filled in toward normal; the report brings it back only through the predictor, and text targets raise the encoder's own lesion readout. We hide each annotated finding on held-out scans whose findings cover a masked cell, with a one-cell margin, and let the predictor fill it in, without and with the patient's report; a lesion probe scores the fill-in (Table~\ref{tab:normal-shift}). All tests were fixed before the results.

\emph{Hidden findings are predicted toward normal (A1).} In every arm, hiding a finding lowers the probe's finding probability (by 0.050 to 0.168) and lesion AUROC by 0.02-0.04. (A1) predicts a shift toward what the context implies, not a loss of the finding. The shift is largest in the conditioned arms ($-$0.124 and $-$0.168): a predictor trained with the report leaves what the context cannot predict to the report, as (A3) expects.

\emph{Conditioning brings the finding back.} Without weak grounding, the patient's report raises the hidden finding's AUROC by 0.044 with conditioning alone and 0.039 with MSP, to the visible level or above. Only with MSP does it move the fill-in toward this lesion more than another patient's report does (own $-$ other in cosine 0.001 against 0.031 [0.025, 0.038]).

\emph{Text targets keep the finding in the encoder (A2).} The encoder's own lesion AUROC is 0.886 without text, 0.892 [0.872, 0.912] with conditioning alone, and 0.959 [0.941, 0.976] with conditioning and MSP. The gap is widest for the 394 findings below 14\,mm that this test excludes: 0.768 [0.731, 0.805] against 0.882 [0.852, 0.907] (Table~\ref{tab:supp-small}). Without the report, the predictor still predicts toward normal in every arm, as (A1) expects: MSP improves the encoder but does not remove this shift.

\subsection{Ablating the report's two roles}
\label{sec:ladder}
\begin{table*}[t]
\centering\small
\caption{Ablating the report at bridge scale. \checkmark: report use kept (Cond.: conditioning; WG: weak grounding); last row: \JECT{}. Plain SigLIP: contrastive training of the same architecture, without latent prediction. Probes: stage-4 token mean; LUNA: 24\,mm decoder cells; R@$K$: image-to-text recall with a pooler trained on the frozen encoder (1,564 scans). $^\dagger$: paired 95\% interval of the difference from \JECT{} excludes zero on both probe sets or in LUNA16 AUROC. $^\ddagger$: no decoder; LUNA reads 24\,mm encoder cells. Bold: best; underlined: second.}
\label{tab:recipe-ladder}
\setlength{\tabcolsep}{5pt}
\begin{tabular}{ccccccccccc}
\toprule
\multicolumn{3}{c}{Report used as} & \multicolumn{2}{c}{Frozen probe} & \multicolumn{2}{c}{LUNA16} & LUNA25 & \multicolumn{3}{c}{Retrieval}\\
\cmidrule(lr){1-3}
Cond. & MSP & WG & Chest-44 & CT-18 & AUROC & CPM & AUROC & R@1 & R@5 & R@10\\
\midrule
\multicolumn{3}{c}{Plain SigLIP~\cite{zhai2023siglip}} & .769$^\dagger$ & .811$^\dagger$ & .775$^{\dagger\ddagger}$ & .108$^\ddagger$ & \textbf{.807}$^\ddagger$ & .058 & .173 & .249\\
\midrule
 \multicolumn{3}{c}{Image-only} & .703$^\dagger$ & .736$^\dagger$ & .682$^\dagger$ & .031 & .692 & .021 & .081 & .123\\
\checkmark & & \checkmark & .757$^\dagger$ & .785$^\dagger$ & .742$^\dagger$ & .076 & .771 & .058 & .194 & .279\\
 & \checkmark & \checkmark & \textbf{.813} & \textbf{.844} & .802 & \underline{.174} & \underline{.785} & \textbf{.162} & \textbf{.366} & \textbf{.474}\\
\checkmark & \checkmark & & .793$^\dagger$ & .830$^\dagger$ & \underline{.803} & .149 & .764 & .099 & .274 & .378\\
\checkmark & \checkmark & \checkmark & \underline{.808} & \underline{.843} & \textbf{.804} & \textbf{.184} & .778 & \underline{.157} & \underline{.352} & \underline{.449}\\
\bottomrule
\end{tabular}
\end{table*}

In our setup, the encoder learns the findings mainly from MSP; conditioning adds little to its frozen readouts. Section~\ref{sec:mechanism} showed that the report repairs the predictor's fill-in. Table~\ref{tab:recipe-ladder} removes one use of the report per run to ask which one teaches the encoder the finding.

\emph{MSP carries the representation (A4).} Removing it lowers the probes by 0.052 and 0.057, LUNA16 cell AUROC by 0.062 and R@1 from 0.157 to 0.058; removing all text lowers the probes by 0.106. A contrastive text target does not replace it: plain SigLIP training of the same architecture falls below \JECT{} on both probes ($-$0.039 and $-$0.032), in cell AUROC ($-$0.029) and in R@1 (0.058 against 0.157), although it is ahead on LUNA25 malignancy ($+$0.029 [0.008, 0.049]).

\emph{Conditioning adds little to the encoder (A3).} Removing it while keeping MSP changes the frozen readouts little (probes $+$0.005 [0.001, 0.008] and $+$0.002 [$-$0.001, 0.004], cell AUROC $-$0.002, R@1 0.162), consistent with the matched pair of Fig.~\ref{fig:clinical-loss}c. LUNA25 changes only when all text is removed ($-$0.086 [$-$0.113, $-$0.060]). We keep conditioning for the predictor: without it the encoder is about as good, but no predictor learns to read the report. The readout of Section~\ref{sec:attribution} is then exactly zero, and a predictor that learns to read the report after pretraining grounds less well than one trained with it from the start.

\subsection{Ablating the architecture}
\label{sec:arch}
\begin{table}[t]
\centering
\caption{Ablating the architecture at bridge scale; columns, $^\dagger$, bold and underline as in Table~\ref{tab:recipe-ladder}.}
\label{tab:arch-ablation}
\resizebox{\columnwidth}{!}{%
\setlength{\tabcolsep}{3pt}
\begin{tabular}{lcccccccc}
\toprule
& \multicolumn{2}{c}{Probe} & \multicolumn{2}{c}{LUNA16} & L25 & \multicolumn{3}{c}{R@$K$}\\
Change & C-44 & CT-18 & AUC & CPM & AUC & 1 & 5 & 10\\
\midrule
None (complete recipe) & \underline{.808} & \underline{.843} & \textbf{.804} & \textbf{.184} & \textbf{.778} & \textbf{.157} & \underline{.352} & \underline{.449}\\
\midrule
No fine-to-coarse connections & .798$^\dagger$ & .830$^\dagger$ & .729$^\dagger$ & .096 & .767 & .106 & .280 & .393\\
Latent prediction at stage 4 only & \textbf{.814} & \textbf{.844} & .781$^\dagger$ & \underline{.155} & \underline{.774} & \underline{.143} & \textbf{.357} & \textbf{.453}\\
No decoder & .802$^\dagger$ & .836$^\dagger$ & \underline{.803} & .145 & .766 & .118 & .290 & .400\\
No latent-prediction losses & .804 & .842 & .781$^\dagger$ & \underline{.155} & .758 & .139 & .320 & .424\\
\bottomrule
\end{tabular}}
\end{table}

The report decides what the encoder learns; the architecture keeps that clinical content at every scale (Table~\ref{tab:arch-ablation}). Fine-to-coarse connections matter most: removing them lowers cell AUROC by 0.075 and the probes by 0.010 and 0.012. Predicting latents at stage 4 alone, or not at all, ties the probes but lowers cell AUROC by 0.023 (separated). Classification thus comes mostly from the report targets; latent prediction adds the spatial readouts, retrieval (R@5 0.352 vs 0.320) and the predictor that the readout reuses (Sec.~\ref{sec:inference}). Removing the decoder lowers the probes by 0.006 and 0.007 (separated) and R@1 from 0.157 to 0.118, with cell AUROC unchanged. The complete recipe is the most balanced: it is best or second best on every readout in Table~\ref{tab:arch-ablation} and on all but LUNA25 in Table~\ref{tab:recipe-ladder}, while each variant that wins one readout loses another (stage-4-only prediction gains 0.006 on Chest-44 but loses 0.023 in cell AUROC).

\subsection{Scaling the recipe}
\label{sec:foundation}
\begin{figure*}[t]
\centering
\includegraphics[width=0.8\textwidth]{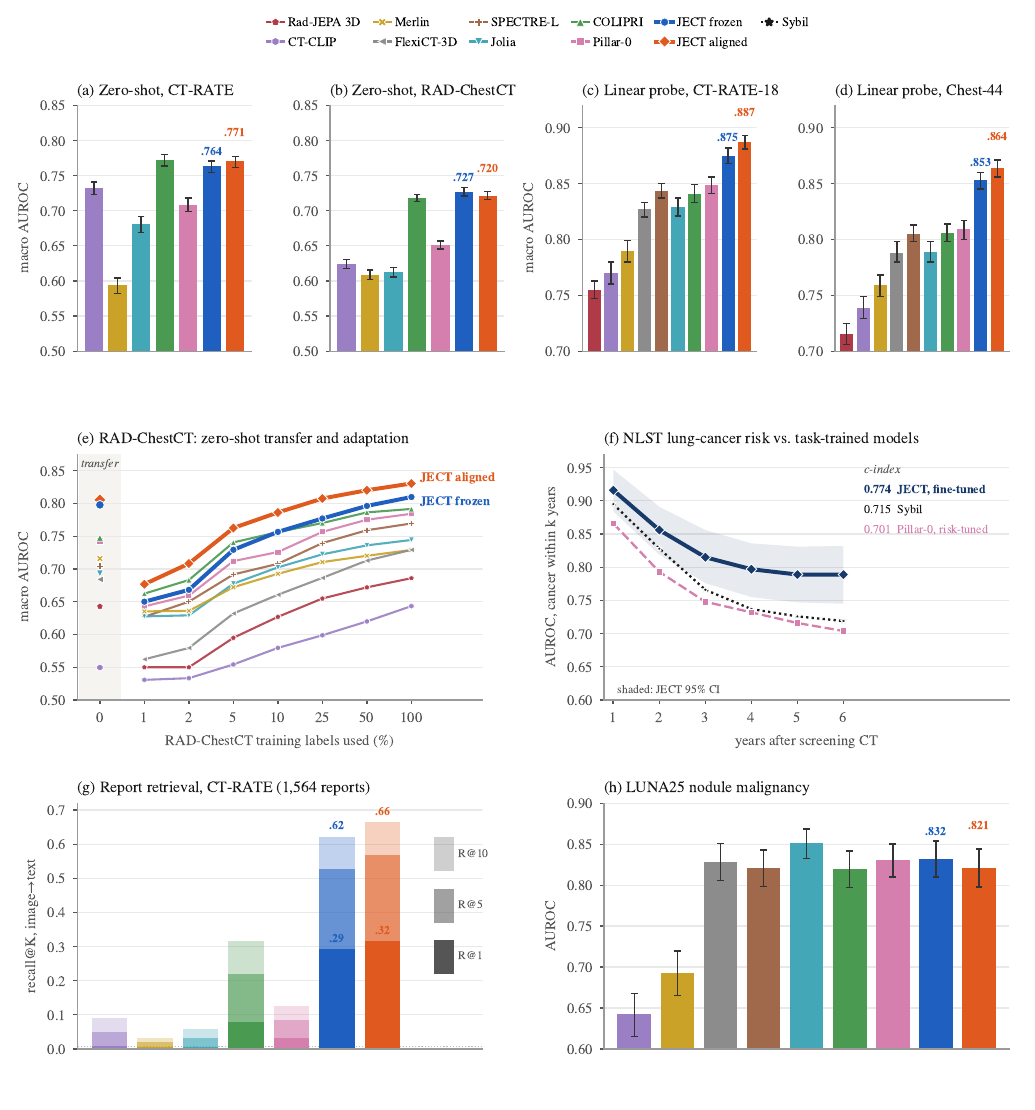}
\caption{\textbf{Foundation-scale comparison with released CT models.} Macro AUROC for (a, b) zero-shot classification and (c, d) linear probes. (e) RAD-ChestCT with a linear head fitted on a fraction of its labels (0\,\%: CT-RATE probes unchanged). (f) NLST lung-cancer AUROC within $k$ years against task-trained models; \JECT{} fine-tuned (shaded: 95\% intervals). (g) Report retrieval on 1,564 CT-RATE scans. (h) LUNA25 malignancy. Released models are frozen. \JECT{} frozen: pretrained encoder, unchanged (a, b, g via a report-aligned pooler); aligned: whole encoder report-aligned from \JECT{}, showing \JECT{} as an initialization. Intervals are in Sec.~\ref{sec:supp-foundation-ci}.}
\label{fig:foundation}
\end{figure*}

Trained at scale, the recipe reaches the level of current CT foundation models. We train it from scratch on the 173,185 foundation volumes (50 epochs, then a 10-epoch cooldown at $320^3$ and 1\,mm) and compare it with eight released CT foundation models (Fig.~\ref{fig:foundation}). Features are read frozen and without the report unless alignment or fine-tuning is stated. Report alignment trains for 15 epochs on the CT-RATE pairs. Our main result, \JECT{} frozen, trains only a pooler on the frozen encoder and leaves every non-text readout unchanged; \JECT{} aligned trains the whole encoder initialized from \JECT{}, showing its value as an initialization for report alignment.

\paragraph{Classification and transfer.}
Without alignment, the frozen 320 encoder reaches 0.875 macro AUROC on CT-RATE-18 and 0.853 on Chest-44, against 0.849 and 0.809 for the best released model, Pillar-0 ($+$0.026 [0.021, 0.031] and $+$0.044 [0.037, 0.051]; CT-RATE-18 labels come from CT-RATE's own report classifier, not from the Chest2Vec model that provides the MSP targets); full alignment adds about 0.01. Rad-JEPA 3D~\cite{trinh2026radjepa3d}, the released CT JEPA, reaches 0.755 and 0.715 from a coarser input (Fig.~\ref{fig:foundation}c, d). On RAD-ChestCT, the frozen encoder transfers the CT-RATE probes unchanged at 0.798, against 0.748 for COLIPRI, but falls slightly below COLIPRI with 1-5\,\% of the labels. Initialized from \JECT{}, the aligned encoder reaches 0.806 ($+$0.058 [0.041, 0.076] over COLIPRI) and stays above the released models at every label fraction (Fig.~\ref{fig:foundation}e). After alignment, zero-shot classification ties COLIPRI (0.771 against 0.772; 0.720 against 0.718), and retrieval ranks the correct report first for 31.5\% of 1,564 CT-RATE scans, against 7.8\% for COLIPRI ($+$0.237 [0.214, 0.262]); with only a pooler on the frozen encoder, \JECT{} frozen reaches 29.2\%.

\paragraph{Lung cancer risk and nodules.}
Fine-tuned on NLST~\cite{nlst2011} against models trained for the task on the same patient resamples, \JECT{} reaches a c-index of 0.774 [0.734, 0.814], above the risk-tuned Pillar-0 ($+$0.073 [0.043, 0.111]) and Sybil~\cite{mikhael2023sybil} ($+$0.059 [0.025, 0.093]). Frozen LUNA25 malignancy (0.832 [0.810, 0.854]) ties the strongest released models, including Jolia ($-$0.019 [$-$0.040, 0.002]). Cross-encoder nodule localization depends on token size (8-24\,mm), so we compare it only within the ablations.

\subsection{The predictor as a report-grounded readout}
\label{sec:attribution}
\begin{table}[t]
\centering\footnotesize
\caption{Deleting report text on 409 held-out ReXGroundingCT cases (733 findings): change in the predictor's error at the finding minus that at lesion-free cells ($\times10^{-3}$; higher is better grounded). Later: the predictor's text pathway trained after pretraining, on the frozen encoder. Own $-$ other: own finding minus another finding of the same patient. Tests were fixed before the results.}
\label{tab:attribution}
\setlength{\tabcolsep}{2.5pt}
\resizebox{\columnwidth}{!}{\begin{tabular}{lcccc}
\toprule
Deleted text & Cond.\ only & MSP + later & Both + later & Both (\JECT{})\\
\midrule
Matching section & 3.7 & 8.6 & 13.7 & \textbf{18.9}\\
Matching sentence & 1.9 & 4.9 & 9.8 & \textbf{15.7}\\
Control sentence & 0.4 & 1.4 & 2.0 & 1.9\\
Own $-$ other finding & 0.6 & 0.6 & 2.2 & \textbf{5.3}\\
\bottomrule
\end{tabular}}
\end{table}

\begin{table}[t]
\centering\footnotesize
\caption{Zero-shot grounding at scale (foundation model before the $320^3$ cooldown; 409 held-out cases, 679 findings with a matched sentence): share of findings whose top cell lies on the finding. No lesion labels in training; brackets: 95\% intervals.}
\label{tab:pointing}
\setlength{\tabcolsep}{4pt}
\begin{tabular}{lc}
\toprule
Map & Pointing (\%)\\
\midrule
Random cell & 4.2\\
COLIPRI, text similarity & 13.3\\
\JECT{}, control-sentence deletion & 11.8\\
\JECT{}, text similarity & 24.8 [20.4, 29.6]\\
\JECT{}, sentence deletion (reused predictor) & \underline{26.6} [21.9, 31.5]\\
\JECT{}, deletion $+$ similarity & \textbf{40.4} [35.4, 45.4]\\
\bottomrule
\end{tabular}
\end{table}

\begin{figure*}[t]
\centering
\includegraphics[width=\textwidth]{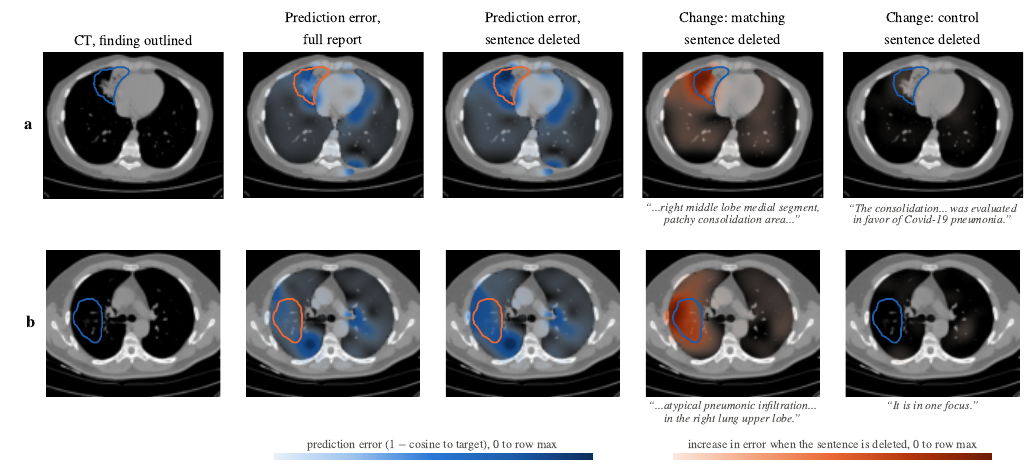}
\caption{\textbf{Deleting one report sentence (foundation model before the $320^3$ cooldown, held-out scans).} Annotated findings (outlined), axial slices in radiological orientation. Columns 2-3: the predictor's error (1 $-$ cosine to the target) with the full report and with the matched sentence deleted. Columns 4-5: increase in error when the matched or a control sentence (quoted) is deleted. 24\,mm cells, smoothed; each row scaled to its maximum. Shown: two of the three findings with the largest matched-minus-control difference; over 532 findings with at least three lesion cells, the matched sentence exceeds its control in 77\%.}
\label{fig:deletion}
\end{figure*}

The predictor trained with both roles can be reused, with no extra training, to show where a report sentence acts in the scan (Sec.~\ref{sec:inference}). After MSP, (A4) predicts that this predictor links report text to places in the image. We test this on 409 held-out ReXGroundingCT cases that no run saw in training. For each annotated finding we delete the sentence that describes it or, as a control, another sentence of the same section, and measure the lesion gap: the change at the finding's cells minus that at lesion-free cells. All tests were fixed before the results (Sec.~\ref{sec:supp-attribution}).

\emph{Deleting a sentence changes the prediction at the finding it describes.} All four tests pass (Table~\ref{tab:attribution}). Deleting the matched sentence raises the error at the finding by 15.7 relative to lesion-free cells ($\times10^{-3}$), eight times the control, and in patients with two findings it changes its own finding more than the other (5.3, separated). Without MSP, the same pathway gives only 1.9, and a direct sentence-to-region map, the model's weak-grounding head, points to the anatomy a sentence names rather than to the lesion (Sec.~\ref{sec:supp-attribution}). Adding text works the same way: a sentence about a nodule in one lobe changes that lobe most in 0.41 of insertions (chance 0.20; 0.81 for consolidation; Fig.~\ref{fig:clinical-loss}b), against 0.21 without MSP.

\emph{Grounding is strongest with both objectives during pretraining.} A predictor trained afterwards on the frozen MSP-only encoder still grounds the report (section-deletion gap 8.6, above the no-MSP arm's 3.7; separated), so MSP supplies the clinical content. With the same later predictor, a model pretrained with conditioning grounds about 60\% more strongly (13.7 against 8.6, $+$5.2 [4.3, 6.1]), although conditioning adds little to its frozen readouts, and training both together reaches 18.9.

\emph{At scale, the readout grounds findings zero-shot.} The foundation model, read at the end of its $224^3$ pretraining (before the $320^3$ cooldown), shows the same gaps (Fig.~\ref{fig:deletion}). Read as zero-shot phrase grounding, with no lesion labels in training (weak grounding sees only organ masks), the reused predictor points to the described finding for 26.6\% of findings, against 11.8\% for the control sentence and 13.3\% for the text map of COLIPRI (Table~\ref{tab:pointing}). The two \JECT{} maps point correctly on different findings, so combining deletion with \JECT{}'s own text-similarity map (through a report-aligned pooler) reaches 40.4\% ($+$0.157 [0.100, 0.213] over the similarity map; Sec.~\ref{sec:supp-grounding}).

\section{Discussion and conclusion}
\label{sec:discussion}
\paragraph{What the objective keeps.}
Latent prediction keeps what the context predicts. The measurements match each argument of Section~\ref{sec:theory}. The argument is not specific to CT: whenever content that matters cannot be predicted from context, latent prediction has no reason to keep it, and a target carrying this content gives that reason back.

\paragraph{What conditioning is for.}
In our setup, conditioning adds little to the encoder, but it trains a predictor that reads the report. With MSP, this predictor can be reused without extra training to show where a report sentence acts in the scan, and pretraining with conditioning makes this report-grounded readout stronger.

\paragraph{Limitations.}
We test only chest CT; extending the recipe to other 3D modalities, such as brain MRI and abdominal CT, is future work. We evaluate the encoder through probes, retrieval and fine-tuning rather than through decoders for detection, segmentation or report generation. This is by design: the context shortcut concerns what the encoder keeps, and a task decoder would mix the encoder's content with what the decoder learns. Building such decoders on \JECT{} features is a natural next step. Our comparisons are within JEPA: we show how clinical text improves a JEPA, not that JEPA beats contrastive pretraining or sets a new state of the art; the uncontrolled foundation comparison only checks that the improvement holds at scale.

\paragraph{Conclusion.}
JECT makes the report both the predictor's condition and the encoder's target, in a hierarchical encoder that keeps clinical content at every scale. The target teaches the encoder the findings the context cannot predict; conditioning trains a predictor that shows where report text acts in the scan.
\section*{LLM Usage Statement}
We used large language models as writing and coding assistants. Claude Opus 5.5 (Anthropic) was used to generate code for training, evaluation, analysis and plotting, to edit the manuscript text, to check the text against the code and the result files. The authors designed the method and the experiments, reviewed and ran all generated code, checked every number and statement against their experiment logs, and take full responsibility for the content.

{\small\bibliographystyle{ieeenat_fullname}\bibliography{main}}

\clearpage
\setcounter{section}{0}
\setcounter{table}{0}
\setcounter{figure}{0}
\setcounter{equation}{0}
\renewcommand{\thesection}{S\arabic{section}}
\renewcommand{\thetable}{S\arabic{table}}
\renewcommand{\thefigure}{S\arabic{figure}}
\renewcommand{\theequation}{S\arabic{equation}}
\renewcommand{\theHsection}{S\arabic{section}}
\renewcommand{\theHtable}{S\arabic{table}}
\renewcommand{\theHfigure}{S\arabic{figure}}
\renewcommand{\theHequation}{S\arabic{equation}}
\maketitlesupplementary
\section{Scope}
This paper studies one failure of JEPA pretraining on clinical images, the
context shortcut, and how to repair it within this family of methods. The
controlled comparisons are between JEPA variants that share the architecture,
data, masking and schedule and differ only in how the report is used. Two
matched controls leave the family: plain SigLIP on the same encoder and pairs
(Section~\ref{sec:supp-ladder}), and a masked autoencoder that differs from the image-only arm
only in its target (Section~\ref{sec:supp-mechanism}). Neither is a ranking of pretraining families.
The foundation-scale comparison is uncontrolled: the released models differ in
data, compute and objective, so it checks that the repair holds at scale
rather than ranking methods.

\section{Method and training details}
\label{sec:supp-implementation}
\label{sec:supp-method}
This section gives the complete method that the main paper summarizes. Unless stated otherwise, every experiment uses this complete recipe.

\begin{figure*}[t]
\centering
\includegraphics[width=\textwidth]{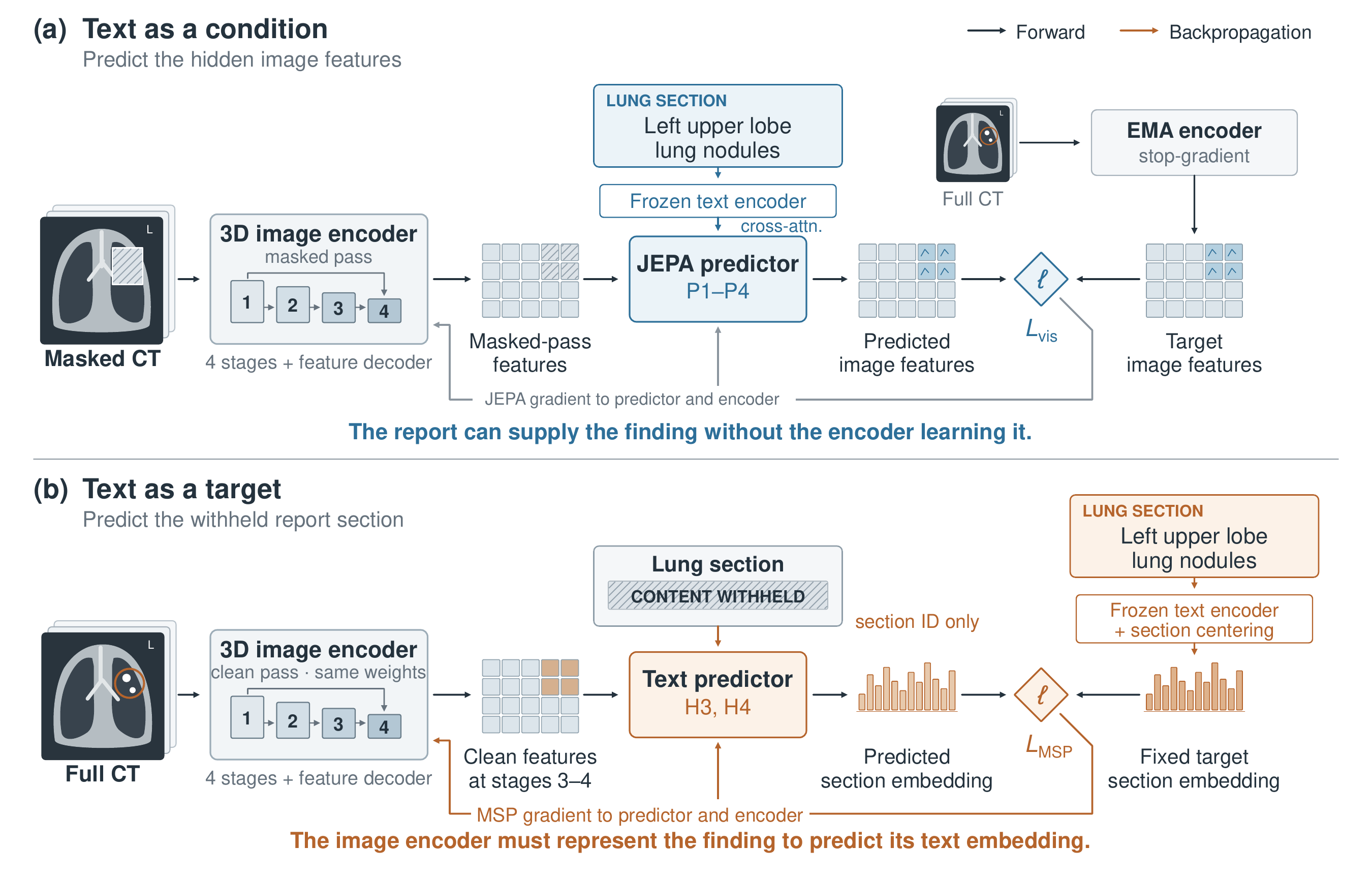}
\caption{\textbf{The report as a condition and as a target.} (a) Conditioning: the predictor reads the report's section embeddings through cross-attention while predicting the hidden image features; the report can supply the finding without the encoder learning it. (b) Masked section-text prediction: on a clean pass, a text predictor that receives only the section's identity predicts the withheld section's embedding from the image, so the encoder must represent the finding.}
\label{fig:supp-roles}
\end{figure*}

\paragraph{Encoder and decoder.}
Let $x$ be a CT volume. A four-stage encoder maps a $224^3$ input to features $e^s$ on grids of $56^3$, $28^3$, $14^3$ and $7^3$ tokens, with channel widths 128, 256, 512 and 1024 and depths 2, 4, 9 and 9. Non-overlapping $4^3$ patch embedding is followed by $2^3$ patch merging between stages; stages 1-3 use Swin-style attention in $7^3$ windows and stage 4 attends globally over its 343 tokens. The encoder has 157.9M parameters, including the fine-to-coarse connections. All visual weights are randomly initialized; the frozen clinical text encoder supplies targets and conditioning but initializes nothing. A top-down decoder refines $e^3$ into $d^3$ with a residual MLP, then upsamples, projects and adds each map to the next finer encoder map ($d^2$ from $d^3$ and $e^2$, $d^1$ from $d^2$ and $e^1$), each followed by a residual MLP and layer normalization. For spatial readouts, $d^3$ and $e^3$ share a lattice but are different transformations. The online branch processes the masked volume; a stop-gradient target encoder and decoder process the unmasked volume, with parameters
\begin{equation}
 \xi\leftarrow m_t\xi+(1-m_t)\theta,
\end{equation}
where $\theta$ denotes the online encoder and decoder parameters. The momentum $m_t$ rises from 0.996 to 0.9999 on a cosine schedule; predictors and text heads are not part of this update.

\paragraph{Fine-to-coarse aggregation.}
Bottom-up connections inject finer features before stages 2-4. Stage 2 receives pooled patch embeddings and stage-1 features; stage 3 additionally receives stage-2 features; stage 4 additionally receives stage-3 features. At each skip entry, each coarse cell queries the finer tokens contained within its spatial extent (the max/mean alternative concatenates local maximum and mean pools instead). For source $f$ and coarse cell $j$, each attention head computes
\begin{equation}
 a_j^{f}=\sum_{i\in\mathcal C_f(j)}
 \operatorname{softmax}_{i}\!\left(
 \frac{q_j^\top k_i}{\sqrt{32}}+b_i\right)v_i,
\end{equation}
where queries derive from the merged coarse feature, keys and values from layer-normalized fine features, and $b_i$ is a learned within-cell 3D-offset bias. Four heads of dimension 32 are concatenated and projected to twice the source width. Patch and stage-1 sources contribute $4^3$ tokens per stage-3 cell; stage-2 features contribute $2^3$. The three pooled sources are concatenated, normalized, projected to 512 channels, and added before the stage-3 blocks. All enclosing skip projections start at zero. Attention is confined to each coarse cell. The final recipe uses this pooler at every skip entry (stages 2, 3 and 4);
the main paper's ablation removes the connections, and Table~\ref{tab:supp-abl-connections} evaluates max/mean pooling and attention at stage 3 only.

\paragraph{Masking and conditioning.}
Coarse cuboids, each side drawn from 4-7 stage-3 cells, are placed until the nominal hidden fraction of 0.485 is reached, with no random fill or buffer; the corresponding patch embeddings are replaced by a learned mask token, and per-stage loss masks identify the hidden targets. Each predictor reads report context through text cross-attention to the 11 cached section embeddings and one sentence sampled from the report (a positive, non-negated sentence 80\% of the time), projected from 2,560 to 512 dimensions. The stage-3 and stage-4 predictors are six-block transformers with 3D rotary positions, each block combining self-attention, text cross-attention and an MLP; the stage-1 and stage-2 predictors are per-token cross-attention heads. Report conditioning is dropped for each sample with probability 0.2, without dropping its supervision targets, and report content never enters the encoder.

\paragraph{Masked and all-token objectives.}
Let $z^s=d_\xi^s(x)$ for $s=1,2,3$, and $z^4=e_\xi^4(x)$. Predictions $p^s$ use the corresponding masked online features and report context. With $\ell(a,b)=1-\cos(a,b)$, the visual objective is
\begin{equation}
\begin{split}
 \mathcal L_{\mathrm{vis}}={}&\sum_{s=1}^{3}\frac{\alpha_s}{|M_s|}
 \sum_{j\in M_s}\ell(p_j^s,\sg[z_j^s])\\
 &+\frac{\alpha_4}{N_4}\sum_{j=1}^{N_4}\ell(p_j^4,\sg[z_j^4]),
\end{split}
\end{equation}
with $\alpha=(0.25,0.35,0.40,0.30)$. Sums include the training batch. Stage 4 predicts all $N_4=7^3$ encoder tokens; ``global'' denotes its attention scope, not a pooled target. Stages 1-3 instead supervise only the masked decoded tokens.

\paragraph{Targets and section identity.}
A frozen Chest2Vec-4B encoder~\cite{ko2026chest2ve} supplies cached 2560-dimensional embeddings for 11 anatomical report sections. We take the 1024-dimensional Matryoshka prefix of each embedding and normalize it to obtain $u_{bk}$ for study $b$ and section $k$; the prefix is a supported truncation of this encoder, not an arbitrary cut. The final recipe masks all valid section contents at the text-head input and predicts them from visual features; absent targets are excluded. Masking here refers to withholding section content from the text head, independently of whether its image input is masked.

Each stage-3/stage-4 text head contains six image cross-attention/MLP blocks. Persistent routing queries combine a learned slot embedding with a projection of the detached training-set section mean $\mu_k$. At every block, this identity is added to the image-derived state for attention queries. Keys and values come exclusively from visual features: decoded $d^3$ at stage 3 and encoder $e^4$ at stage 4. The initial state is zero and the first cross-attention has no query residual, so section identity routes attention without being added directly to the output state. There is no attention between section queries. Before output projection, a section-dependent gate
\begin{equation}
 g_k=1+0.5\tanh\!\left(\operatorname{LN}(W\mu_k+v_k)\right)
\end{equation}
scales the image-derived state elementwise, where $v_k$ is the slot embedding. Section means are population statistics, not the patient's target embedding. The final heads use no anatomical-coordinate bias.

\paragraph{Clinical-innovation regression.}
To reduce reliance on the section prior, form residual targets
\begin{equation}
 \rho_{bk}=u_{bk}-a_t\mu_k,\quad
 w_{bk}=\|\rho_{bk}\|_2,\quad
 \widetilde u_{bk}=\frac{\rho_{bk}}{w_{bk}+\epsilon}.
\end{equation}
The centering factor $a_t$ ramps from zero to one; the routing queries always use the full mean. If $\widehat u_{bk}^s$ is the predicted embedding and $v_{bk}^{\mathrm{valid}}$ indicates an eligible target, then
\begin{equation}
 \mathcal L_{\mathrm{MSP}}=
 \frac12\sum_{s\in\{3,4\}}\gamma_s
 \frac{\sum_{b,k}v_{bk}^{\mathrm{valid}}w_{bk}
 \ell(\widehat u_{bk}^s,\widetilde u_{bk})}
 {\sum_{b,k}v_{bk}^{\mathrm{valid}}w_{bk}+\epsilon},
\end{equation}
with $\gamma_3=0.5$ and $\gamma_4=1$. Normalization and weights are computed from detached targets. This loss regresses continuous clinical embeddings without batch negatives or token generation.

\paragraph{Clean-pass schedule.}
The complete recipe samples $b_t\sim\operatorname{Bernoulli}(0.5)$ once per optimizer window, shared across devices and accumulated microbatches. When $b_t=1$, an additional unmasked pass through the shared online encoder and decoder feeds MSP and weak grounding; when $b_t=0$, both heads are skipped and masked prediction continues with its own report dropout. The gate also requires valid supervision, and no inverse-probability correction rescales the active losses.

\paragraph{Weak grounding.}
On active clean-pass windows, a separate grounding head uses cached finding-sentence embeddings as queries over stage-2 decoded features. It predicts maps for their referenced anatomical regions using weak region masks, rather than lesion-segmentation targets. Sentence queries cross-attend to image tokens; projected query-token dot products produce spatial logits. For logits $l$ and binary target $A$, define
\begin{equation}
\begin{split}
 B(l,A)={}&\tfrac12\operatorname{BCE}(l,A)\\
 &+\tfrac12\left(1-
 \frac{2\sum_j\sigma(l_j)A_j+1}
 {\sum_j\sigma(l_j)+\sum_j A_j+1}\right).
\end{split}
\end{equation}
We average $B$ over valid sentence-region pairs and add half-weighted supervision of a learned null query against the union of valid referenced regions. This defines $\mathcal L_{\mathrm{WG}}$. The grounding slots are separate from the 11 MSP section slots, with up to six valid regions per study.

\paragraph{Dense prediction and regularization.}
A dense auxiliary objective supervises visible stage-2 and stage-3 tokens using text-free predictor calls. It minimizes channel-averaged squared distance between layer-normalized predictions and detached targets. Spatial weights decrease as $1,2/3,1/3$ across three successive visible rings around the mask and are zero elsewhere. The weighted loss is normalized by the sum of spatial weights. Stage weights are $0.35\times0.15$ and $0.40\times0.10$, respectively; their sum defines $\mathcal L_{\mathrm{dense}}$.

SIGReg~\cite{balestriero2025lejepa} regularizes separate 256-dimensional projections of online features, averaged across the four stages. Each stage uses at most 512 sampled tokens and 512 random unit projection directions. The Epps-Pulley statistic compares projected empirical characteristic functions with a standard Gaussian, using 17 integration knots over $[0,3]$ and symmetry-weighted trapezoidal integration. Variance and covariance statistics are monitored but are not additional penalties in the SIGReg-only recipe.

\paragraph{Combined objective and optimization.}
The complete recipe uses the training loss
\begin{equation}
\begin{split}
 \mathcal L={}&\mathcal L_{\mathrm{vis}}+
 d_t\mathcal L_{\mathrm{dense}}+0.02\mathcal L_{\mathrm{SIGReg}}\\
 &+b_t\tau_t\left(0.5\mathcal L_{\mathrm{MSP}}+
 0.5\mathcal L_{\mathrm{WG}}\right).
\end{split}
\end{equation}
For fractional training progress $q_t$, $\tau_t=\min(1,q_t/f_\tau)$ and $a_t=\min(1,q_t/f_a)$, with $f_\tau=0.01$ and $f_a=0.2$ at bridge scale and $f_\tau=0.001356$ at foundation scale; at bridge scale $d_t$ is zero through 10\% of training and rises linearly to one by 40\%. We use AdamW (learning rate $7\times10^{-5}$, weight decay 0.05), bfloat16 arithmetic, gradient clipping at norm 1 and an effective batch of 32: one device with microbatch 8 and four-step accumulation at bridge scale, two devices with two-step accumulation at foundation scale. Gradients update the online branch and the active heads; the EMA targets and the clinical text encoder receive none.

\paragraph{Reproducibility.} The learning rate warms up linearly for one epoch and then follows a cosine decay to 1\% of its base value; AdamW uses $\beta=(0.9,0.999)$. A bridge run is 36,700 optimizer steps (733 per epoch), over which the EMA momentum schedule spans. Intensity augmentation shifts HU by up to $\pm$20, scales by up to 3\% and adds Gaussian noise (s.d.\ 0.01); and the volume is random cropped to 224 from 256. A fixed per-step seed (20260829) fixes masks, report dropout and the clean-pass draw across arms. \emph{Alignment} trains a pooler and projection on the frozen encoder with a sigmoid loss for 15 epochs at batch 96 (AdamW, learning rate $10^{-4}$, weight decay 0.05, 300 warm-up steps); the Chest2Vec text tower is adapted with low-rank adapters (rank 16, $\alpha=32$, learning rate $5\times10^{-5}$), and the logit bias starts at $-10$. \emph{Linear probes} are one linear layer on frozen features, trained with class-balanced BCE for 50 epochs (Adam, learning rate $10^{-3}$, weight decay $10^{-4}$, batch 512, cosine decay), five seeds (41-45) ensembled.

\section{Evaluation protocols}
\label{sec:supp-eval}
\subsection{Statistics}
\label{sec:supp-stats}
Ablation comparisons are paired: we resample patients (3,000 bootstrap draws; 2,000 for attribution) and report 95\% intervals. The foundation-scale tables state their own resampling unit and number of draws in their captions. A difference is \emph{separated} when its interval excludes zero; for an ablation we require this on both probe label sets or on LUNA16 cell AUROC. Each ablation is a single pretraining run, so intervals reflect the test data; second seeds of three arms (Section~\ref{sec:supp-ladder}) show that retraining moves the probes by under 0.004. Cell CPM is not compared across foundation models, because it depends on each encoder's token size. Intervals not given in the main text are in the tables of this supplement.

\subsection{Classification}
Linear probes use the 1,544-scan CT-RATE validation cohort, with macro AUROC over the chest2vec-44 and CT-RATE-18 label sets and five probe seeds. The complete-recipe ladder and the foundation model read the stage-4 token mean. Patient-cluster bootstrap intervals capture evaluation-cohort uncertainty, not the variation from retraining the visual model.

\subsection{LUNA spatial evaluation}
\label{sec:supp-luna}
LUNA16 uses 622 public scans; LUNA25 uses the public set, 6,062 nodules in 4,003 scans from 2,103 patients after preprocessing exclusions. Lung-centred $384^3$ cubes at 1\,mm are resized to $256^3$ and centre-cropped to $224^3$ (336\,mm field of view), with intensities $(\mathrm{HU}+1000)/2000$.
\emph{Localization:} class-balanced logistic regression ($C=1$, five scan-grouped folds) on 24\,mm stage-3 decoder cells. A lung cell is positive when it contains a nodule's centre or at least 10\% of it lies inside the nodule, a sphere of the annotated diameter (radius at least 2\,mm), so every nodule has a positive cell; the nodules are small (median diameter 6.4\,mm), and 777 of 848 are positive only through their centre cell. A one-cell ring around positives is excluded, and nodule-free scans count toward false positives. We report cell AUROC and cell CPM, the mean sensitivity at 0.125-8 false-positive cells per scan; this is not the LUNA16 challenge protocol.
\emph{Malignancy:} class-balanced logistic regression ($C=0.5$, five patient-grouped folds) on the nodule's centre token concatenated with the mean over its $3^3$ neighbourhood.

\subsection{Alignment and retrieval}
Alignment trains a pooler and projection on each frozen encoder for 15 epochs on the 23,476 CT-RATE training pairs. Retrieval uses the 1,564-scan validation gallery in both directions; its 1,503 distinct-report groups let duplicate reports count as correct.

\subsection{Data and licenses}
\label{sec:supp-data}
All external data are public and de-identified: CT-RATE~\cite{hamamci2026ctrate} (bridge pretraining, probes, retrieval), ReXGroundingCT~\cite{baharoon2025rexgroundingct} (finding masks for the hidden-finding and readout tests), LUNA16~\cite{setio2017luna16} from LIDC-IDRI~\cite{armato2011lidc} via TCIA~\cite{clark2013tcia}, LUNA25~\cite{peeters2026luna25} from the National Lung Screening Trial~\cite{nlst2011} (CC BY 4.0), RAD-ChestCT~\cite{draelos2021radchest}, and TotalSegmentator~\cite{wasserthal2023totalsegmentator} masks for lobes and regions. Each is used under its terms, and no images are redistributed. The institutional volumes, used only at foundation scale, were de-identified before training.
Table~\ref{tab:supp-splits} lists every dataset, its split and its use. No evaluation scan or patient is in the pretraining data of the model it evaluates. The Chest2Vec text encoders, which supply the MSP targets and the retrieval text tower, were trained without the CT-RATE validation reports, so no evaluation report reaches a text model during training.

\begin{table*}[t]
\centering\footnotesize
\caption{Datasets and splits. Counts are volumes unless stated. Folds are grouped by patient (LUNA16: by scan) so that no patient appears in both training and test data of a readout.}
\label{tab:supp-splits}
\setlength{\tabcolsep}{4pt}
\begin{tabular}{@{}p{2.6cm}p{3.0cm}p{3.6cm}p{7.4cm}@{}}
\toprule
Dataset & Split & Size & Use\\
\midrule
CT-RATE & train & 23,476 & bridge pretraining; alignment pairs; probe training (23,475 CT-RATE-18, 23,206 Chest-44)\\
CT-RATE & validation & 1,564 (1,544 shared probe cohort) & probes, zero-shot, retrieval gallery (1,503 report groups)\\
ReXGroundingCT & CT-RATE validation & 409 cases, 733 findings & hidden-finding test, report attribution (held out from all runs' reports)\\
Institutional (first CT per patient, 2003-2020) + CT-RATE train & pretraining & 173,185 & foundation pretraining only\\
RAD-ChestCT & all mapped & 3,630 (3,629 for \JECT{}) & zero-shot transfer\\
RAD-ChestCT & official train / test & 2,286 (2,114 patients) / 360 & label efficiency (fixed patient subsets of train, scored on test)\\
LUNA16 & 5 scan folds & 622 scans, 848 nodules & localization (frozen readout, out-of-fold)\\
LUNA25 & 5 patient folds & 6,062 nodules, 4,003 scans, 2,103 patients & malignancy (frozen readout, out-of-fold)\\
NLST & Sybil split, train / dev / test & 5,533 / 1,330 / 1,309 exams & lung-cancer risk fine-tuning; checkpoint chosen on dev\\
\bottomrule
\end{tabular}
\end{table*}

\paragraph{Ethics.} Use of the institutional data, used only for foundation pretraining, was approved by the institutional review board of the contributing hospital; all scans and reports were de-identified before training. The public datasets, including CT-RATE and RAD-ChestCT, were obtained and used under their respective data use agreements and terms of use.

\section{Testing the context shortcut}
\label{sec:supp-mechanism}
\paragraph{Protocol.} A stage-3 cell is positive when at least 10\% of its voxels belong to an annotated lesion; findings without such a cell are excluded. Of the 1,127 findings, 733 are eligible; the excluded findings are small (median equivalent-sphere diameter 10\,mm), and none below 14\,mm is eligible. This test therefore measures larger findings; the LUNA16 localization readout covers small nodules. The visible condition keeps the lesion and a one-cell ring; the hidden condition hides both, with other mask locations adjusted to keep the hidden fraction fixed. With the image hidden, the predictor receives no report, the patient's report, or another patient's report. A standardized logistic probe is fitted on unmasked target features (five folds holding out lesion and normal patients, eight position-matched normal cells per positive cell) and applied to each model's fill-in; each model has its own probe. The probe is binary: it separates cells of any annotated finding from normal cells, so it measures whether a model represents an abnormality, not which finding it is.

\paragraph{Small findings.} The 394 findings excluded from the main test (no 24\,mm cell with 10\% lesion; all below 14\,mm) were scored with their highest-coverage cell, as in LUNA16 (Table~\ref{tab:supp-small}). Image-only pretraining represents them worst: its encoder's lesion AUROC falls from 0.886 on the main findings to 0.768, while the complete recipe keeps 0.920. A finding this small fills under a tenth of its cell, so hiding it barely changes the fill-in at this resolution.

\begin{table}[t]
\centering\tiny
\caption{Encoder lesion AUROC (any finding vs normal cells) on the held-out cases, for the 733 findings of the main test and the 394 smaller ones it excludes. Brackets: 95\% patient-bootstrap intervals.}
\label{tab:supp-small}
\setlength{\tabcolsep}{3pt}
\begin{tabular}{lccccc}
\toprule
Arm & Cond. & MSP & WG & Main (733) & Small (394)\\
\midrule
Image-only & & & & .886 [.860, .910] & .768 [.731, .805]\\
$+$Conditioning & \checkmark & & & .892 [.872, .912] & .782 [.749, .823]\\
$+$MSP & & \checkmark & & .967 [.952, .980] & .892 [.863, .919]\\
$+$Conditioning, MSP & \checkmark & \checkmark & & .959 [.941, .976] & .882 [.852, .907]\\
$+$MSP, WG & & \checkmark & \checkmark & .965 [.948, .979] & .913 [.888, .938]\\
$+$(\JECT{}) & \checkmark & \checkmark & \checkmark & .965 [.949, .979] & .920 [.897, .942]\\
\bottomrule
\end{tabular}
\end{table}

\paragraph{Gradient statement of (A2).} Let $\theta$ be the trainable parameters, $\hat z_h=g_\theta(f_\theta(v))$ the prediction, and $z_h$ the target, held fixed by the stop-gradient. The squared error $\lVert\hat z_h\rVert^2-2\hat z_h^\top z_h+\lVert z_h\rVert^2$ and the cosine loss $1-\hat z_h^\top z_h/(\lVert\hat z_h\rVert\lVert z_h\rVert)$ are linear in $z_h$ (in $z_h/\lVert z_h\rVert$ for the cosine), apart from a term without $\theta$. Since $\hat z_h$ depends on $v$ alone, $\mathbb E[\nabla_\theta\mathcal L\mid v]$ is the same loss gradient evaluated at $\mathbb E[z_h\mid v]$. Writing $z_h=\mathbb E[z_h\mid v]+\varepsilon$, the part $\varepsilon$ that $v$ cannot predict contributes zero expected gradient to every parameter, encoder included. This concerns one hidden region under one mask; a finding visible under another mask can still be learned when it helps predict hidden content there.

\paragraph{Encoder lesion signal during training.} We repeated the image-only arm, saving a checkpoint every five epochs, and measured the encoder's lesion AUROC on the held-out cases at each one. It rises from 0.744 at epoch 5 to 0.864 at epoch 30 and then stops: 0.878 at both epochs 45 and 50 (final run of Table~\ref{tab:normal-shift}: 0.886). Training longer does not raise this ceiling, while the same architecture, data and schedule with MSP alone reach 0.967. The limit comes from the objective, as (A2) predicts: without a text target, the encoder gets no signal to learn what the context cannot predict.

\paragraph{A fixed pixel target.} Fig.~\ref{fig:mechanism} allows that image-only JEPA keeps
less finding signal than a pixel target does. For pixel and feature targets alike, the error left at a hidden finding does not teach the encoder the finding (A2). A JEPA target, however, is computed by the encoder itself, so the objective can also lower that error by representing the finding less; a pixel target is fixed and cannot be lowered this way. We tested this with a masked autoencoder that differs from the image-only arm only in its target: normalized voxel patches instead of EMA features, with the same encoder, predictor, masking, data and schedule. The autoencoder is ahead on the probes (0.742 and 0.772 against 0.703 and 0.736; $+$0.039 [0.032, 0.046] and $+$0.036 [0.029, 0.042]) and in encoder lesion AUROC on the small findings (0.803 against 0.768; $+$0.036 [0.003, 0.069]); on the main findings the difference is not separated (0.903 against 0.886; $+$0.017 [$-$0.004, 0.037]). This is consistent with a fixed target keeping more of the finding, but the training curve above shows a ceiling, not a decline, so we do not claim that the JEPA target actively suppresses findings. A fixed target is not enough, however: it still gives no signal for what the context cannot predict, and the complete recipe exceeds the autoencoder by 0.062 [0.044, 0.081] and 0.117 [0.091, 0.145] in lesion AUROC. Repairing the shortcut takes a target that carries the finding, which the report provides.

\paragraph{MSP alone.} Trained without conditioning or weak grounding (Table~\ref{tab:supp-target-only}), MSP alone raises the encoder's lesion AUROC from 0.886 to 0.967 on the main findings and from 0.768 to 0.892 on the small ones. This matches the complete recipe on the main findings ($-$0.002 [$-$0.013, 0.008]); the complete recipe adds most on the small findings ($+$0.028 [0.005, 0.052]) and on the probes. Without latent prediction, MSP alone loses 0.042 on the main findings and 0.087 on the small ones, so the JEPA objective still contributes when the report is a target.

\paragraph{Weak grounding complements MSP.} Weak grounding is a second report target, not a form of conditioning: it runs on the same clean pass as MSP and asks the encoder to map the region each finding sentence names, while the report never enters the predictor through it. Added to MSP without conditioning (Table~\ref{tab:supp-target-only}), it raises the probes by 0.018 [0.013, 0.023] and 0.012 [0.009, 0.016], the small-finding lesion AUROC from 0.892 to 0.913, and R@1 from 0.109 to 0.162, while the main-finding lesion AUROC is unchanged (0.965 against 0.967). Conditioning adds little to the encoder on top of these targets: with MSP alone the main-finding lesion AUROC is 0.967 without it and 0.959 with it, and with MSP and weak grounding 0.965 in both cases. The encoder gains of the complete recipe therefore come from its two report targets, MSP first, with weak grounding adding most for small findings and retrieval; conditioning acts at the predictor.

\begin{table}[t]
\centering\small
\caption{Report targets without conditioning: MSP alone and with weak grounding. Probe: macro AUROC (stage-4 mean); Lesion: encoder lesion AUROC on the held-out main (733) and small (394) findings; R@1: frozen-encoder image-to-text retrieval.}
\label{tab:supp-target-only}
\setlength{\tabcolsep}{3pt}
\resizebox{\columnwidth}{!}{%
\begin{tabular}{lccccc}
\toprule
& \multicolumn{2}{c}{Probe} & \multicolumn{2}{c}{Lesion} & \\
Arm & C-44 & CT-18 & 733 & 394 & R@1\\
\midrule
Image-only & .703 & .736 & .886 & .768 & .021\\
MSP alone & .794 & .832 & .967 & .892 & .109\\
MSP alone, no latent prediction & .781 & .819 & .925 & .805 & .045\\
MSP $+$ weak grounding & .813 & .844 & .965 & .913 & .162\\
\JECT{} & .808 & .843 & .965 & .920 & .157\\
\bottomrule
\end{tabular}}
\end{table}

\section{Ablations}
\label{sec:supp-ladder}
\paragraph{Training.}
Every arm trains on the 23,476 CT-RATE volumes for 50 epochs on one device (microbatch 8, four-step accumulation). A fixed per-step seed makes mask sampling, report dropout and the clean-pass draw identical across arms. Each arm changes one setting: MSP weight 0 (no MSP); report dropout 1 (no conditioning); all text off (image-only); weak-grounding weight 0; no fine-to-coarse connections; max/mean pooling; attention at stage 3 only; clean-pass probability 0.25 or 1; or all latent-prediction weights 0. The no-decoder arm passes the layer-normalized encoder stages $e^1$-$e^3$ through in place of the decoder. Plain SigLIP~\cite{zhai2023siglip} trains the same encoder from scratch with only a sigmoid contrastive loss on the same pairs (50 epochs, Chest2Vec-4B text tower with low-rank adapters); without a decoder, its LUNA readouts and the complete recipe's comparison use stage-3 encoder tokens.

\paragraph{Reading the variant tables.}
Tables~\ref{tab:supp-abl-target}-\ref{tab:supp-abl-connections} use the protocol and columns of the main paper's ablation table; recalls at 5 and 10 are in Table~\ref{tab:supp-ladder-retrieval}. Each table repeats the complete recipe as its first row. $^\dagger$: the paired 95\% interval of the difference from the complete recipe excludes zero on both probe label sets (probe columns) or in LUNA16 cell AUROC.

\newcommand{\ablhead}{%
\toprule
& \multicolumn{2}{c}{Probe} & \multicolumn{2}{c}{LUNA16} & LUNA25 & \\
Arm & C-44 & CT-18 & AUC & CPM & AUC & R@1\\
\midrule}

\paragraph{Report target granularity.}
Table~\ref{tab:supp-abl-target} replaces section targets with the whole report or single utterances; both also drop the section routing query and centre globally, so a fourth arm removes only those. That arm ties the complete recipe (R@1 0.178), so the routing query does not drive the result. Whole-report targets lower the probes (0.011, 0.008) and R@1 (0.134); utterance targets perform like sections (Chest-44 $+$0.008 [0.003, 0.014], R@1 0.173). MSP needs targets finer than the whole report; we keep sections, which follow the report's structure.

\begin{table}[t]
\centering\footnotesize
\caption{Report target granularity: what each MSP target covers. $^\ast$: no section routing query, global centring. $^\S$: weak grounding uses the 11 utterance slots instead of 6. C-44: Chest-44; AUC: AUROC.}
\label{tab:supp-abl-target}
\setlength{\tabcolsep}{3pt}
\setlength{\tabcolsep}{2.5pt}\begin{tabular}{lcccccc}
\ablhead
Section (complete) & .808 & .843 & .804 & .184 & .778 & .157\\
Whole report$^\ast$ & .798$^\dagger$ & .834$^\dagger$ & .811 & .165 & .783 & .134\\
Utterance$^{\ast\S}$ & .817 & .844 & .803 & .161 & .776 & .173\\
Section$^\ast$ & .809 & .844 & .806 & .177 & .772 & .178\\
\bottomrule
\end{tabular}
\end{table}

\paragraph{MSP loss form.}
Regressing the plain section embedding, uncentred and unweighted, matches the complete recipe on the frozen readouts (Table~\ref{tab:supp-abl-loss}; only Chest-44 separates, $+$0.005 [0.001, 0.010]) and grounds the report about as well (Table~\ref{tab:supp-msp-loss}). The gain does not depend on the loss form.

\begin{table}[t]
\centering\footnotesize
\caption{MSP loss form: frozen readouts.}
\label{tab:supp-abl-loss}
\setlength{\tabcolsep}{3pt}
\setlength{\tabcolsep}{2.5pt}\begin{tabular}{lcccccc}
\ablhead
Centred (complete) & .808 & .843 & .804 & .184 & .778 & .157\\
Plain cosine & .813 & .844 & .795 & .142 & .793 & .164\\
\bottomrule
\end{tabular}
\end{table}

\begin{table}[t]
\centering\footnotesize
\caption{MSP loss form: report grounding of the predictor on the tests of the main paper. Deletion: mean change in fill residual at the lesion when its sentence is deleted, minus the complete recipe (negative = weaker). Lobe: fraction of 402 scans where an inserted ``nodule'' or ``consolidation'' sentence changes its named lobe most (chance 0.2).}
\label{tab:supp-msp-loss}
\setlength{\tabcolsep}{3pt}
\begin{tabular}{lccc}
\toprule
Arm & Deletion & Lobe: nodule & Lobe: consol. \\
\midrule
Complete recipe  & --         & .41 & .81 \\
Plain-cosine MSP & $-$.0033   & .42 & .76 \\
No MSP           & $-$.0139   & .21 & .40 \\
\bottomrule
\end{tabular}
\end{table}

\paragraph{When the text target is applied.}
The text heads read a clean pass that runs in a random share $p$ of optimizer windows. Lowering $p$ to 0.25 costs 0.014 and 0.015 on the probes and 0.017 [0.002, 0.032] in cell AUROC (Table~\ref{tab:supp-abl-schedule}). Raising $p$ to 1 gains 0.017 [0.012, 0.022] and 0.010 [0.007, 0.014] on the probes, 0.023 [0.008, 0.038] in cell AUROC and 0.045 in R@1, and keeps the predictor's report grounding: the lobe tests give 0.43 and 0.77
(complete recipe: 0.41, 0.81). The encoder improves the more often the text target is applied. We fixed $p=0.5$ before these experiments because each clean pass adds a forward and backward pass (26\% more training time per step at $p=1$), and keep it for every reported model; the main results are therefore conservative for this recipe.

\begin{table}[t]
\centering\footnotesize
\caption{When the text target is applied: clean-pass probability $p$, and the pass the text heads read.}
\label{tab:supp-abl-schedule}
\setlength{\tabcolsep}{3pt}
\setlength{\tabcolsep}{2.5pt}\begin{tabular}{lcccccc}
\ablhead
$p=0.25$ & .794$^\dagger$ & .828$^\dagger$ & .786$^\dagger$ & .153 & .783 & .107\\
$p=0.5$ & .808 & .843 & .804 & .184 & .778 & .157\\
$p=1$ & .825$^\dagger$ & .853$^\dagger$ & .826$^\dagger$ & .227 & .786 & .202\\
Masked pass & .785$^\dagger$ & .816$^\dagger$ & .761$^\dagger$ & .105 & .769 & .079\\
\bottomrule
\end{tabular}
\end{table}

\paragraph{Latent-prediction loss weights.}
Equal stage weights ($\alpha_s=0.25$), or scoring stage 4 only on half-masked cells, do not separate from the complete recipe (probes within 0.003; cell AUROC $-$0.013 [$-$0.027, 0.002] and $-$0.008 [$-$0.023, 0.007]; Table~\ref{tab:supp-abl-weights}).

\begin{table}[t]
\centering\footnotesize
\caption{Latent-prediction loss weights and scored cells.}
\label{tab:supp-abl-weights}
\setlength{\tabcolsep}{3pt}
\setlength{\tabcolsep}{2.5pt}\resizebox{\columnwidth}{!}{%
\begin{tabular}{lcccccc}
\ablhead
Complete recipe & .808 & .843 & .804 & .184 & .778 & .157\\
Equal, $\alpha_s=0.25$ & .806 & .841 & .791 & .163 & .784 & .154\\
Stage 4, masked cells & .808 & .843 & .795 & .159 & .788 & .142\\
\bottomrule
\end{tabular}}
\end{table}

\paragraph{Fine-to-coarse connections.}
Keeping the connections but changing their pooling matters less than removing them: max/mean pooling is within 0.004 on the probes and 0.007 in cell AUROC, while attention at stage 3 alone is 0.017 [0.004, 0.030] lower in cell AUROC (Table~\ref{tab:supp-abl-connections}).

\begin{table}[t]
\centering\footnotesize
\caption{Fine-to-coarse connections: pooling rule.}
\label{tab:supp-abl-connections}
\setlength{\tabcolsep}{3pt}
\setlength{\tabcolsep}{2.5pt}\resizebox{\columnwidth}{!}{%
\begin{tabular}{lcccccc}
\ablhead
Attention (complete) & .808 & .843 & .804 & .184 & .778 & .157\\
Max/mean, all & .805 & .839 & .796 & .131 & .781 & .145\\
Attention, stage 3 & .807 & .842 & .787$^\dagger$ & .137 & .790 & .136\\
\bottomrule
\end{tabular}}
\end{table}

\paragraph{Second seed.} A second pretraining seed of the complete recipe gives probes of 0.811 and 0.842 (seed 1: 0.808, 0.843), LUNA16 cell AUROC 0.798 (0.804), CPM 0.153 (0.184), LUNA25 0.793 (0.778) and R@1 0.162 (0.157). Probe and cell-AUROC differences between seeds are within about 0.006, while CPM moves by 0.031; the main text therefore draws spatial conclusions from cell AUROC. Second seeds of the image-only arm and of conditioning alone without weak grounding agree as closely: probes of 0.702 and 0.738 (seed 1: 0.703, 0.736) and of 0.698 and 0.731 (seed 1: 0.695, 0.732). Seed variation is thus far below the 0.10-0.11 gaps between these arms and the complete recipe.

\paragraph{Readouts}
Probes read the stage-4 token mean (the stage-3 mean is 0.012 and 0.017 lower for the complete recipe). Table~\ref{tab:supp-ladder-retrieval} lists all recalls.

\begin{table}[t]
\centering\footnotesize
\caption{Frozen-encoder retrieval for the complete-recipe ladder (1,564 scans, 1,503 report groups).}
\label{tab:supp-ladder-retrieval}
\setlength{\tabcolsep}{3pt}
\resizebox{\columnwidth}{!}{%
\begin{tabular}{lcccccc}
\toprule
& \multicolumn{3}{c}{Image to text} & \multicolumn{3}{c}{Text to image}\\
Encoder & R@1 & R@5 & R@10 & R@1 & R@5 & R@10\\
\midrule
Complete recipe & .157 & .352 & .449 & .169 & .381 & .474\\
Image-only & .021 & .081 & .123 & .028 & .086 & .134\\
No MSP & .058 & .194 & .279 & .072 & .207 & .306\\
No conditioning & .162 & .366 & .474 & .192 & .384 & .491\\
No WG & .099 & .274 & .378 & .113 & .311 & .418\\
No connections & .106 & .280 & .393 & .133 & .332 & .442\\
Max/mean pooling & .145 & .343 & .437 & .181 & .375 & .474\\
Attention at stage 3 & .136 & .333 & .424 & .152 & .367 & .460\\
Clean-pass 0.25 & .107 & .301 & .393 & .137 & .315 & .428\\
Clean-pass 1 & .202 & .414 & .520 & .214 & .460 & .559\\
Masked pass & .079 & .234 & .332 & .095 & .279 & .380\\
No latent prediction & .139 & .320 & .424 & .132 & .340 & .446\\
Stage-4 prediction only & .143 & .357 & .453 & .177 & .384 & .481\\
Plain-cosine MSP & .164 & .384 & .488 & .184 & .396 & .525\\
Whole-report MSP & .134 & .318 & .418 & .127 & .327 & .439\\
Utterance MSP & .173 & .371 & .474 & .189 & .393 & .506\\
Section, no routing & .178 & .374 & .473 & .194 & .391 & .505\\
Equal stage weights & .154 & .359 & .468 & .176 & .376 & .472\\
Stage 4, masked cells & .142 & .353 & .462 & .166 & .376 & .491\\
\bottomrule
\end{tabular}}
\end{table}

\paragraph{Claim boundaries.}
The matched experiments support adding MSP to conditioned latent prediction and the spatial benefit of the fine-to-coarse connections. The gradient statement of (A2) holds for one masked loss with a stop-gradient target; we do not derive the combined training dynamics with EMA targets, visible-token losses and SIGReg, and we compare with a pixel target only as a matched control (Section~\ref{sec:supp-mechanism}), not across pretraining families.

\section{Foundation scale}
\label{sec:supp-foundation}
\paragraph{Training data.}
The foundation model trains the complete recipe (Section~\ref{sec:supp-method}) from random
initialization on 173,185 volumes: the first chest CT of each patient at one tertiary hospital (2003-2020), with section-structured reports and de-identified before training, plus the CT-RATE training studies. The institutional validation and test patients, the CT-RATE validation studies, RAD-ChestCT, LUNA16 and LUNA25 are disjoint from this set. It trains for 50 epochs at $224^3$ and 1.5\,mm on two devices (effective batch 32), then cools down for 10 epochs at $320^3$ and 1\,mm with learning rate $2.5\times10^{-5}$; the main paper evaluates the model after the cooldown.

\paragraph{RAD-ChestCT protocol.}
All 3,629 shared scans are converted from the released arrays with their true geometry before extraction. For label efficiency (main paper, Fig.~\ref{fig:foundation}e), a linear head per encoder is fitted on fixed patient subsets of the official training split and scored on the official test split over the 16 labels shared with CT-RATE-18, five patient subsets per fraction, three probe seeds ensembled per subset, with shared patient-cluster bootstrap draws; at 0\% the CT-RATE-fitted probes are applied unchanged.

\subsection{Resolution ablation}
\label{sec:supp-res-ablation}
\begin{table*}[t]
\centering\footnotesize
\caption{The foundation encoder before (224: epoch 50 at $224^3$) and after the 10-epoch $320^3$, 1\,mm cooldown (320), each read with the protocol of the foundation tables (Tables~\ref{tab:supp-fnd-probe}--\ref{tab:supp-fnd-nlst-task}). LUNA16 has no cross-model table and uses the protocol of Section~\ref{sec:supp-luna} on stage-3 decoder cells: 24\,mm at $224^3$ and 16\,mm at $320^3$. Probes: frozen stage-4 token mean, paired patient bootstrap (3,000 draws) of the difference. Other rows give each model's own 95\% interval. R@1: frozen encoder with a report-aligned pooler (frozen), and whole encoder aligned (aligned).}
\label{tab:supp-res-ablation}
\setlength{\tabcolsep}{6pt}
\begin{tabular}{lccc}
\toprule
Readout & 224 & 320 & 320 $-$ 224\\
\midrule
Probe, Chest-44 & .848 & .853 & $+$.005 [.001, .010]\\
Probe, CT-RATE-18 & .872 & .875 & $+$.004 [.001, .007]\\
Zero-shot, CT-RATE-18 & .765 [.756, .773] & .764 [.755, .771] & \\
R@1, frozen & .272 [.250, .294] & .292 [.269, .316] & \\
R@1, aligned & .284 [.262, .308] & .315 [.293, .338] & \\
LUNA16 CPM & .257 [.235, .280] & .413 [.373, .459] & \\
LUNA16 cell AUROC & .911 [.898, .923] & .936 [.923, .948] & \\
LUNA25 malignancy & .824 [.801, .846] & .832 [.810, .854] & \\
NLST c-index, fine-tuned & .757 [.714, .797] & .774 [.734, .814] & \\
\bottomrule
\end{tabular}
\end{table*}

Table~\ref{tab:supp-res-ablation} compares the foundation encoder before and after the $320^3$ cooldown. The cooldown raises the probes ($+$0.004-0.005), R@1 (0.272 to 0.292 frozen, 0.284 to 0.315 aligned), LUNA16 cell AUROC (0.911 to 0.936) and CPM (0.257 to 0.413), and the fine-tuned NLST c-index (0.757 to 0.774); zero-shot is unchanged. Much of the CPM gain is token size: read from its 12\,mm stage-2 tokens, the 224 encoder already reaches 0.408 [0.379, 0.439].

\subsection{Results with confidence intervals}
\label{sec:supp-foundation-ci}
Exact values behind the main paper's foundation figure, with 95\% bootstrap intervals. CT-RATE results are in-domain for \JECT{}, whose pretraining includes the CT-RATE training studies (not the validation studies); RAD-ChestCT, LUNA25 and NLST are external; bold marks the best result and every result whose interval overlaps it. \JECT{} rows are the 320 model: frozen, the pretrained encoder (with a report-aligned pooler where text is needed); aligned, the whole encoder report-aligned from \JECT{}. FlexiCT-3D~\cite{flexi3d} and SPECTRE-L have no usable text tower, so their zero-shot and retrieval cells are blank. Rad-JEPA 3D, a released image-only CT JEPA, has no text tower either and appears in the probe and RAD-ChestCT tables; it is read with the same protocol (mean of its 1,024 final tokens) from its own input, a $32\times256\times256$ resize of the body region (about 8-12\,mm between slices), which is coarser than the other models' inputs.

\begin{table}[t]
\centering\scriptsize
\caption{Frozen linear probes on CT-RATE validation, macro AUROC (patient-cluster bootstrap, 3,000 draws). \JECT{} frozen: the pretrained 320 encoder, no alignment (stage-4 token mean); aligned: the pooled feature of the whole encoder after report alignment. FlexiCT-3D, SPECTRE-L and Rad-JEPA 3D intervals come from a separate bootstrap and are not paired with the other rows.}
\label{tab:supp-fnd-probe}
\setlength{\tabcolsep}{3pt}
\begin{tabular}{lcc}
\toprule
Model & Chest-44 & CT-RATE-18\\
\midrule
Rad-JEPA 3D & .715\,{\color{gray}\tiny[.706, .725]} & .755\,{\color{gray}\tiny[.747, .763]}\\
CT-CLIP & .739\,{\color{gray}\tiny[.729, .749]} & .770\,{\color{gray}\tiny[.760, .780]}\\
Merlin & .759\,{\color{gray}\tiny[.749, .768]} & .790\,{\color{gray}\tiny[.780, .799]}\\
FlexiCT-3D & .788\,{\color{gray}\tiny[.780, .798]} & .827\,{\color{gray}\tiny[.820, .833]}\\
SPECTRE-L & .805\,{\color{gray}\tiny[.798, .813]} & .843\,{\color{gray}\tiny[.837, .850]}\\
Jolia & .789\,{\color{gray}\tiny[.780, .798]} & .829\,{\color{gray}\tiny[.821, .837]}\\
COLIPRI & .806\,{\color{gray}\tiny[.798, .814]} & .841\,{\color{gray}\tiny[.833, .849]}\\
Pillar-0 & .809\,{\color{gray}\tiny[.800, .817]} & .849\,{\color{gray}\tiny[.841, .856]}\\
\midrule
JECT frozen & \textbf{.853}\,{\color{gray}\tiny[.845, .860]} & \textbf{.875}\,{\color{gray}\tiny[.868, .882]}\\
JECT aligned & \textbf{.864}\,{\color{gray}\tiny[.856, .871]} & \textbf{.887}\,{\color{gray}\tiny[.881, .893]}\\
\bottomrule
\end{tabular}
\end{table}

\begin{table*}[t]
\centering\scriptsize
\caption{Zero-shot classification with text prompts (``\{\} is present. / ``\{\} is not present.), macro AUROC; CT-RATE validation (1,564 / 1,544 scans) and all 3,630 RAD-ChestCT scans with mapped labels (3,629 for \JECT{}, one scan without a cube; scan bootstrap, 100 draws).}
\label{tab:supp-fnd-zeroshot}
\setlength{\tabcolsep}{3pt}
\begin{tabular}{lcccc}
\toprule
Model & CT-RATE-18, CT-RATE & CT-RATE-18, RAD-ChestCT & Chest-44, CT-RATE & Chest-44, RAD-ChestCT\\
\midrule
CT-CLIP & .732\,{\color{gray}\tiny[.723, .741]} & .625\,{\color{gray}\tiny[.617, .631]} & .674\,{\color{gray}\tiny[.663, .687]} & .593\,{\color{gray}\tiny[.586, .600]}\\
Merlin & .594\,{\color{gray}\tiny[.582, .604]} & .609\,{\color{gray}\tiny[.602, .616]} & .545\,{\color{gray}\tiny[.536, .556]} & .563\,{\color{gray}\tiny[.556, .570]}\\
Jolia & .681\,{\color{gray}\tiny[.669, .692]} & .613\,{\color{gray}\tiny[.606, .619]} & .644\,{\color{gray}\tiny[.634, .653]} & .583\,{\color{gray}\tiny[.576, .589]}\\
COLIPRI & \textbf{.772}\,{\color{gray}\tiny[.764, .780]} & \textbf{.718}\,{\color{gray}\tiny[.713, .723]} & .739\,{\color{gray}\tiny[.730, .749]} & \textbf{.703}\,{\color{gray}\tiny[.696, .710]}\\
Pillar-0 & .708\,{\color{gray}\tiny[.699, .718]} & .652\,{\color{gray}\tiny[.645, .657]} & .662\,{\color{gray}\tiny[.652, .674]} & .623\,{\color{gray}\tiny[.616, .629]}\\
\midrule
JECT frozen & \textbf{.764}\,{\color{gray}\tiny[.755, .771]} & \textbf{.727}\,{\color{gray}\tiny[.721, .733]} & \textbf{.750}\,{\color{gray}\tiny[.740, .758]} & \textbf{.691}\,{\color{gray}\tiny[.685, .698]}\\
JECT aligned & \textbf{.771}\,{\color{gray}\tiny[.762, .778]} & \textbf{.720}\,{\color{gray}\tiny[.716, .727]} & \textbf{.757}\,{\color{gray}\tiny[.750, .766]} & \textbf{.693}\,{\color{gray}\tiny[.687, .700]}\\
\bottomrule
\end{tabular}
\end{table*}

\begin{table*}[t]
\centering\scriptsize
\caption{Report retrieval on CT-RATE validation (1,564 images, 1,503 unique reports; duplicate reports count as correct), each model with its own image and text towers (query bootstrap, 1,000 draws).}
\label{tab:supp-fnd-retrieval}
\setlength{\tabcolsep}{3pt}
\begin{tabular}{lcccccc}
\toprule
Model & I$\to$T R@1 & I$\to$T R@5 & I$\to$T R@10 & T$\to$I R@1 & T$\to$I R@5 & T$\to$I R@10\\
\midrule
CT-CLIP & .010\,{\color{gray}\tiny[.005, .015]} & .050\,{\color{gray}\tiny[.040, .061]} & .090\,{\color{gray}\tiny[.076, .104]} & .015\,{\color{gray}\tiny[.009, .021]} & .053\,{\color{gray}\tiny[.042, .063]} & .089\,{\color{gray}\tiny[.076, .102]}\\
Merlin & .004\,{\color{gray}\tiny[.001, .008]} & .022\,{\color{gray}\tiny[.015, .029]} & .033\,{\color{gray}\tiny[.024, .042]} & .007\,{\color{gray}\tiny[.003, .011]} & .028\,{\color{gray}\tiny[.020, .037]} & .041\,{\color{gray}\tiny[.031, .051]}\\
Jolia & .006\,{\color{gray}\tiny[.003, .010]} & .033\,{\color{gray}\tiny[.025, .043]} & .058\,{\color{gray}\tiny[.046, .068]} & .009\,{\color{gray}\tiny[.004, .013]} & .025\,{\color{gray}\tiny[.018, .033]} & .045\,{\color{gray}\tiny[.035, .057]}\\
COLIPRI & .078\,{\color{gray}\tiny[.065, .091]} & .219\,{\color{gray}\tiny[.199, .238]} & .316\,{\color{gray}\tiny[.293, .339]} & .120\,{\color{gray}\tiny[.104, .137]} & .287\,{\color{gray}\tiny[.265, .310]} & .373\,{\color{gray}\tiny[.350, .397]}\\
Pillar-0 & .031\,{\color{gray}\tiny[.023, .040]} & .084\,{\color{gray}\tiny[.071, .099]} & .127\,{\color{gray}\tiny[.109, .144]} & .021\,{\color{gray}\tiny[.015, .029]} & .060\,{\color{gray}\tiny[.049, .072]} & .093\,{\color{gray}\tiny[.079, .108]}\\
\midrule
JECT frozen & \textbf{.292}\,{\color{gray}\tiny[.269, .316]} & \textbf{.526}\,{\color{gray}\tiny[.504, .554]} & \textbf{.622}\,{\color{gray}\tiny[.600, .647]} & \textbf{.317}\,{\color{gray}\tiny[.292, .341]} & \textbf{.543}\,{\color{gray}\tiny[.518, .569]} & \textbf{.639}\,{\color{gray}\tiny[.615, .663]}\\
JECT aligned & \textbf{.315}\,{\color{gray}\tiny[.293, .338]} & \textbf{.569}\,{\color{gray}\tiny[.547, .593]} & \textbf{.664}\,{\color{gray}\tiny[.642, .688]} & \textbf{.343}\,{\color{gray}\tiny[.319, .367]} & \textbf{.585}\,{\color{gray}\tiny[.560, .609]} & \textbf{.677}\,{\color{gray}\tiny[.653, .702]}\\
\bottomrule
\end{tabular}
\end{table*}

\begin{table*}[t]
\centering\scriptsize
\caption{RAD-ChestCT label efficiency: macro AUROC over 16 labels on the 360 official test scans when a linear head is fitted on the given fraction of the training split (five fixed patient subsets per fraction, three probe seeds ensembled per subset; 0\,\% = CT-RATE-18 probes applied unchanged; patient bootstrap, 1,000 draws).}
\label{tab:supp-fnd-radchest}
\setlength{\tabcolsep}{3pt}
\begin{tabular}{lcccccccc}
\toprule
Model & 0\,\% & 1\,\% & 2\,\% & 5\,\% & 10\,\% & 25\,\% & 50\,\% & 100\,\%\\
\midrule
Rad-JEPA 3D & .643\,{\color{gray}\tiny[.620, .668]} & .550\,{\color{gray}\tiny[.535, .562]} & .550\,{\color{gray}\tiny[.536, .562]} & .595\,{\color{gray}\tiny[.579, .610]} & .627\,{\color{gray}\tiny[.611, .644]} & .655\,{\color{gray}\tiny[.637, .676]} & .672\,{\color{gray}\tiny[.653, .693]} & .686\,{\color{gray}\tiny[.663, .708]}\\
CT-CLIP & .550\,{\color{gray}\tiny[.522, .576]} & .531\,{\color{gray}\tiny[.519, .543]} & .533\,{\color{gray}\tiny[.519, .547]} & .554\,{\color{gray}\tiny[.539, .570]} & .580\,{\color{gray}\tiny[.563, .596]} & .599\,{\color{gray}\tiny[.579, .618]} & .620\,{\color{gray}\tiny[.600, .640]} & .643\,{\color{gray}\tiny[.618, .667]}\\
Merlin & .716\,{\color{gray}\tiny[.692, .739]} & .635\,{\color{gray}\tiny[.620, .650]} & .636\,{\color{gray}\tiny[.621, .650]} & .672\,{\color{gray}\tiny[.654, .689]} & .693\,{\color{gray}\tiny[.672, .712]} & .711\,{\color{gray}\tiny[.691, .730]} & .720\,{\color{gray}\tiny[.702, .740]} & .729\,{\color{gray}\tiny[.709, .750]}\\
FlexiCT-3D & .684\,{\color{gray}\tiny[.662, .708]} & .562\,{\color{gray}\tiny[.549, .575]} & .580\,{\color{gray}\tiny[.566, .593]} & .632\,{\color{gray}\tiny[.616, .648]} & .661\,{\color{gray}\tiny[.644, .679]} & .686\,{\color{gray}\tiny[.666, .707]} & .713\,{\color{gray}\tiny[.694, .733]} & .729\,{\color{gray}\tiny[.707, .751]}\\
SPECTRE-L & .704\,{\color{gray}\tiny[.678, .729]} & .627\,{\color{gray}\tiny[.611, .641]} & .650\,{\color{gray}\tiny[.636, .663]} & .692\,{\color{gray}\tiny[.673, .709]} & .708\,{\color{gray}\tiny[.688, .726]} & .740\,{\color{gray}\tiny[.721, .758]} & .759\,{\color{gray}\tiny[.740, .776]} & .769\,{\color{gray}\tiny[.750, .787]}\\
Jolia & .693\,{\color{gray}\tiny[.668, .715]} & .628\,{\color{gray}\tiny[.614, .642]} & .629\,{\color{gray}\tiny[.616, .643]} & .678\,{\color{gray}\tiny[.662, .695]} & .703\,{\color{gray}\tiny[.685, .722]} & .722\,{\color{gray}\tiny[.703, .744]} & .736\,{\color{gray}\tiny[.715, .758]} & .744\,{\color{gray}\tiny[.721, .768]}\\
COLIPRI & .748\,{\color{gray}\tiny[.727, .768]} & \textbf{.662}\,{\color{gray}\tiny[.648, .677]} & \textbf{.683}\,{\color{gray}\tiny[.671, .696]} & \textbf{.741}\,{\color{gray}\tiny[.724, .758]} & \textbf{.756}\,{\color{gray}\tiny[.738, .773]} & .770\,{\color{gray}\tiny[.751, .789]} & \textbf{.786}\,{\color{gray}\tiny[.768, .805]} & .792\,{\color{gray}\tiny[.773, .811]}\\
Pillar-0 & .743\,{\color{gray}\tiny[.723, .764]} & .643\,{\color{gray}\tiny[.628, .658]} & .659\,{\color{gray}\tiny[.646, .674]} & .712\,{\color{gray}\tiny[.696, .728]} & .726\,{\color{gray}\tiny[.710, .743]} & .757\,{\color{gray}\tiny[.739, .775]} & .775\,{\color{gray}\tiny[.757, .793]} & .784\,{\color{gray}\tiny[.766, .801]}\\
\midrule
JECT frozen & \textbf{.798}\,{\color{gray}\tiny[.779, .815]} & .650\,{\color{gray}\tiny[.638, .662]} & .668\,{\color{gray}\tiny[.654, .682]} & .729\,{\color{gray}\tiny[.713, .745]} & \textbf{.757}\,{\color{gray}\tiny[.740, .772]} & \textbf{.777}\,{\color{gray}\tiny[.759, .794]} & \textbf{.796}\,{\color{gray}\tiny[.778, .813]} & \textbf{.810}\,{\color{gray}\tiny[.789, .828]}\\
JECT aligned & \textbf{.806}\,{\color{gray}\tiny[.786, .825]} & \textbf{.677}\,{\color{gray}\tiny[.662, .690]} & \textbf{.708}\,{\color{gray}\tiny[.695, .722]} & \textbf{.763}\,{\color{gray}\tiny[.747, .778]} & \textbf{.786}\,{\color{gray}\tiny[.772, .801]} & \textbf{.808}\,{\color{gray}\tiny[.789, .825]} & \textbf{.820}\,{\color{gray}\tiny[.804, .837]} & \textbf{.831}\,{\color{gray}\tiny[.815, .846]}\\
\bottomrule
\end{tabular}
\end{table*}

\begin{table}[t]
\centering\scriptsize
\caption{LUNA25 malignancy of 6,062 nodules from each model's local feature at the nodule, AUROC (five patient folds, patient bootstrap, 3,000 draws). \JECT{} frozen reads stage-3 decoder cells, as in the ablations; the aligned encoder, which has no decoder, reads stage-3 encoder tokens (0.825 for the frozen encoder read the same way).}
\label{tab:supp-fnd-luna}
\setlength{\tabcolsep}{3pt}
\begin{tabular}{lc}
\toprule
Model & LUNA25 AUROC\\
\midrule
CT-CLIP & .642\,{\color{gray}\tiny[.615, .668]}\\
Merlin & .693\,{\color{gray}\tiny[.665, .720]}\\
FlexiCT-3D & \textbf{.829}\,{\color{gray}\tiny[.806, .851]}\\
SPECTRE-L & \textbf{.821}\,{\color{gray}\tiny[.798, .843]}\\
Jolia & \textbf{.851}\,{\color{gray}\tiny[.833, .868]}\\
COLIPRI & \textbf{.820}\,{\color{gray}\tiny[.797, .841]}\\
Pillar-0 & \textbf{.830}\,{\color{gray}\tiny[.810, .850]}\\
\midrule
JECT frozen & \textbf{.832}\,{\color{gray}\tiny[.810, .854]}\\
JECT aligned & \textbf{.821}\,{\color{gray}\tiny[.798, .844]}\\
\bottomrule
\end{tabular}
\end{table}

\begin{table*}[t]
\centering\scriptsize
\caption{NLST lung-cancer risk against models trained for the task, scored with the comparison package on identical patient resamples (1,309 test scans, follow-up from cancer-free days; AUC: 1,000 resamples, c-index: 200). Sybil: released ensemble; Pillar0-Sybil-1.5: Pillar-0 fine-tuned for risk; JECT fine-tuned: whole encoder with a Sybil-style head, 12 epochs, checkpoint chosen by development c-index; JECT 224 is the encoder before the $320^3$ cooldown.}
\label{tab:supp-fnd-nlst-task}
\setlength{\tabcolsep}{3pt}
\begin{tabular}{lccccccc}
\toprule
Model & 1 y & 2 y & 3 y & 4 y & 5 y & 6 y & c-index\\
\midrule
Sybil & \textbf{.895}\,{\color{gray}\tiny[.855, .932]} & \textbf{.828}\,{\color{gray}\tiny[.786, .869]} & \textbf{.766}\,{\color{gray}\tiny[.717, .814]} & \textbf{.737}\,{\color{gray}\tiny[.684, .787]} & \textbf{.726}\,{\color{gray}\tiny[.672, .773]} & \textbf{.719}\,{\color{gray}\tiny[.666, .770]} & \textbf{.715}\,{\color{gray}\tiny[.667, .760]}\\
Pillar0-Sybil-1.5 & \textbf{.866}\,{\color{gray}\tiny[.824, .908]} & \textbf{.793}\,{\color{gray}\tiny[.751, .837]} & \textbf{.748}\,{\color{gray}\tiny[.702, .798]} & \textbf{.732}\,{\color{gray}\tiny[.686, .782]} & \textbf{.716}\,{\color{gray}\tiny[.669, .765]} & \textbf{.704}\,{\color{gray}\tiny[.654, .754]} & \textbf{.701}\,{\color{gray}\tiny[.657, .746]}\\
\midrule
JECT 224, fine-tuned & \textbf{.899}\,{\color{gray}\tiny[.864, .934]} & \textbf{.845}\,{\color{gray}\tiny[.806, .887]} & \textbf{.801}\,{\color{gray}\tiny[.759, .844]} & \textbf{.777}\,{\color{gray}\tiny[.731, .820]} & \textbf{.767}\,{\color{gray}\tiny[.720, .811]} & \textbf{.771}\,{\color{gray}\tiny[.724, .816]} & \textbf{.757}\,{\color{gray}\tiny[.714, .797]}\\
JECT 320, fine-tuned & \textbf{.916}\,{\color{gray}\tiny[.884, .947]} & \textbf{.856}\,{\color{gray}\tiny[.819, .891]} & \textbf{.815}\,{\color{gray}\tiny[.776, .856]} & \textbf{.797}\,{\color{gray}\tiny[.755, .836]} & \textbf{.789}\,{\color{gray}\tiny[.747, .831]} & \textbf{.789}\,{\color{gray}\tiny[.745, .832]} & \textbf{.774}\,{\color{gray}\tiny[.734, .814]}\\
\bottomrule
\end{tabular}
\end{table*}

\section{Report-grounded readout}
\subsection{Attribution protocol}
\label{sec:supp-attribution}
The design and the four tests were fixed before any final result. The residual is $1-\cos$ between the filled latent and the EMA target at each stage-3 cell, averaged over eight seeded masks shared across report conditions and models; the change is $A=R(r_\text{edited})-R(r)$. The holder section is the section most similar to a finding, the matched sentence the most similar sentence, and the control the least similar sentence of the same section; deleting a sentence re-embeds its section without it. The lesion gap is the mean $A$ over the finding's masked cells (at least 10\% lesion) minus the mean over lesion-free body cells. The tests are a positive holder gap (F1), a matched sentence above its control (F2), and both above the no-MSP arm (F3, F4). The which-lesion contrast uses patients with two findings matched to different sentences.

The cohort is the 409 ReXGroundingCT cases on CT-RATE validation scans with a sectioned report: 733 findings, 679 with sentence statistics, 104 patients for which-lesion; intervals use 2,000 shared patient draws. Without conditioning the text pathway is untrained, so the change is exactly zero.

\paragraph{Readout details.} As a localizer the readout is modest (same-anatomy AUROC 0.702; 0.554 without MSP), but the weak-grounding head does no better within an anatomical region (0.520). The later predictor is trained with the pretraining loss and report dropout; its image pathway was itself pretrained with or without the report, so the comparison isolates pretraining with conditioning, not the encoder alone. Joint training adds a further 5.2 (18.9 against 13.7), partly because the joint predictor trains for 50 epochs. In the insertion test, swapping right and left moves the change to the other side, a negated sentence localizes near chance, and a predictor trained afterwards places nodules above chance only on the complete-recipe encoder (0.43).

\paragraph{Direct-map baseline.} We compare the readout with the arm's own weak-grounding head, a direct text-to-region map queried with each edited sentence and pooled to the same cells. The head names the lobe or side almost perfectly, as it was trained to, and wins the which-lesion contrast, whose two findings usually lie in different regions. Within a region it is at its null level (same-anatomy AUROC 0.520, against 0.702 for the readout) and ignores negation; the readout carries both, and loses both without MSP (0.554).

\subsection{Zero-shot phrase grounding}
\label{sec:supp-grounding}
\paragraph{Protocol.} The cases, findings, sentences and cells are those of the attribution analysis. Both of \JECT{}'s maps come from one checkpoint, the foundation model at the end of its $224^3$ pretraining, before the $320^3$ cooldown. The deletion map is the increase in the predictor's residual when the finding's matched sentence is deleted. The similarity map is the cosine between the sentence embedding and each 48\,mm stage-4 token passed alone through a report-aligned pooler trained on this frozen checkpoint (maximum over its 11 latents). COLIPRI and Merlin are read through their released dense paths, from our 1\,mm volumes through their official preprocessing, at the same cell centres.

\paragraph{Scores.} Pointing counts a finding as hit when the map's highest cell contains any of its voxels (a random cell: 4.2\%). Region is the median number of cells at or above half the map's maximum, and precision the share of them on the finding (a finding covers 7 cells). Same-region AUROC separates the finding's cells from lesion-free cells of the same region. The combined map is the mean of the two maps' within-scan ranks, fixed before scoring.

\begin{table*}[t]
\centering\small
\caption{Zero-shot phrase grounding on the 409 held-out cases (679 findings). Pointing: share of findings whose highest cell lies on the finding (random cell 0.042). Region: median cells at or above half the map's maximum (a finding covers 7). Brackets: 95\% patient-bootstrap intervals (2,000 draws). The lower block tests the combination: the sham replaces the deletion map with the deletion map of the control sentence; the second similarity map is plain SigLIP's attention map.}
\label{tab:supp-grounding}
\setlength{\tabcolsep}{4pt}
\begin{tabular}{@{}lcccc@{}}
\toprule
Map & Pointing & Region & Precision & Same-region AUROC\\
\midrule
\multicolumn{5}{@{}l}{\emph{Foundation model, one checkpoint}}\\
Sentence deletion & .266 [.219, .315] & 8 & .231 & .696\\
Control-sentence deletion & .118 & - & - & -\\
Text similarity (readout head) & .248 [.204, .296] & 131 & .142 & .702\\
Deletion + similarity (rank mean) & \textbf{.404} [.354, .454] & - & - & .715\\
\midrule
\multicolumn{5}{@{}l}{\emph{Alignment head on bridge encoders}}\\
With MSP (complete recipe) & .215 & 263 & .079 & .697\\
Image-only JEPA & .067 & 317 & .056 & .528\\
Plain SigLIP, dense & .178 & 270 & .074 & .676\\
Plain SigLIP, attention & .226 & 66 & .185 & .692\\
\midrule
\multicolumn{5}{@{}l}{\emph{Released models}}\\
COLIPRI & .133 & 364 & .059 & .477\\
Merlin & .060 & 330 & .048 & .482\\
\midrule
\multicolumn{5}{@{}l}{\emph{Controls for the combination (pointing; paired differences)}}\\
Sham deletion + similarity & .250 & \multicolumn{3}{l}{combination $-$ sham: $+$.154 [.110, .201]}\\
Two similarity maps & .250 & \multicolumn{3}{l}{combination $-$ two similarity maps: $+$.155 [.095, .210]}\\
Combination $-$ similarity & & \multicolumn{3}{l}{$+$.157 [.100, .213]; combination $-$ deletion: $+$.139 [.092, .186]}\\
\bottomrule
\end{tabular}
\end{table*}

\paragraph{Findings.} Within a finding's region, the deletion and similarity maps rank lesion cells equally well (0.696 and 0.702) and point equally often ($+$0.018 [$-$0.044, 0.076]), but the deletion map marks a finding-sized region (8 cells) where the similarity map covers the named region (131). They rarely peak on the same finding (rank correlation 0.08), so their combination reaches 0.404; neither a sham deletion map nor a second similarity map gives this gain. The same alignment head points to findings on the bridge encoder with MSP (0.215) and near chance on the image-only encoder (0.067). COLIPRI is above chance (0.133) but does not rank lesion cells within their region.

\begin{figure*}[t]
\centering
\includegraphics[width=\textwidth]{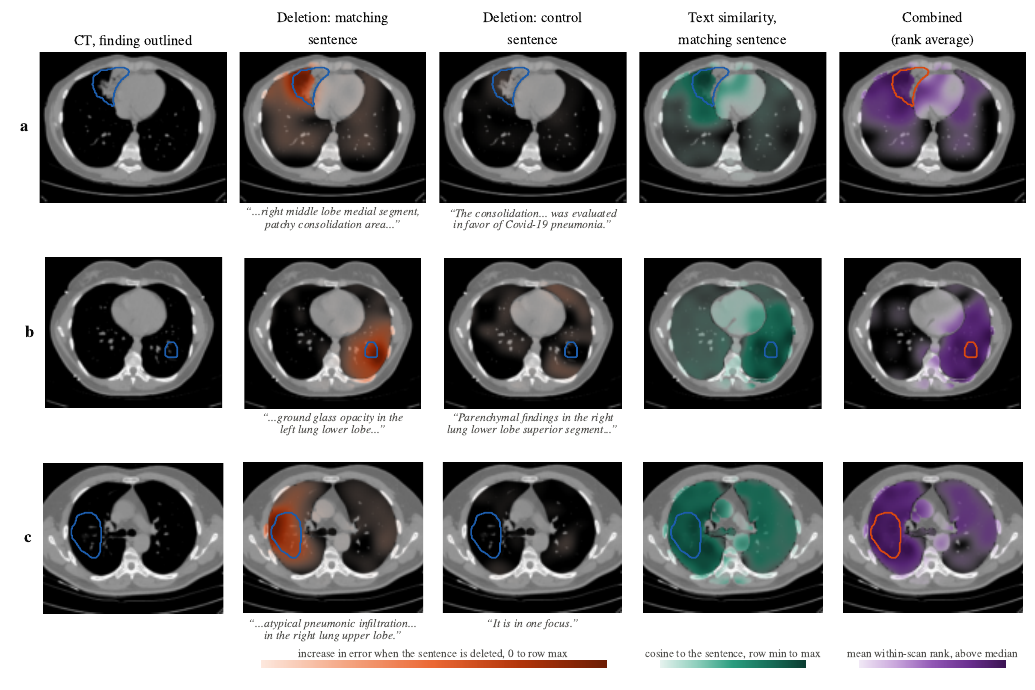}
\caption{Findings with the largest deletion gaps, chosen before the combination was computed, with all of the foundation model's maps. Deletion maps: increase in the predictor's error when the matching or control sentence is deleted. Text similarity: cosine between the matching sentence and each 48\,mm token through that pooler, from the row's minimum to maximum. Combined: mean within-scan rank of the two, shown above the median. The similarity map covers the region the sentence names; the deletion map marks a few cells within it. The combination's gain is in where its single highest cell falls (Table~\ref{tab:supp-grounding}), not in a sharper map.}
\label{fig:supp-grounding}
\end{figure*}

\end{document}